\documentclass[lettersize,journal]{IEEEtran}
\ifCLASSOPTIONcompsoc
  \usepackage[nocompress]{cite}
\else
  \usepackage{cite}
\fi
\usepackage[T1]{fontenc}    % use 8-bit T1 fonts
\usepackage{url}            % simple URL typesetting
\usepackage{booktabs}       % professional-quality tables
\usepackage{amsfonts}       % blackboard math symbols
\usepackage{nicefrac}       % compact symbols for 1/2, etc.
\usepackage{microtype}      % microtypography
\usepackage{xcolor}         % colors
\usepackage{graphicx}
\usepackage{algorithmicx}
\usepackage{algorithm}
\usepackage{amsmath}
\usepackage{amsthm}
\usepackage{amssymb}
\usepackage{ifthen}
\usepackage{algpseudocode}
\usepackage{multirow}
\usepackage{ragged2e}
\usepackage{bbding}

\newcommand{\partitle}[1]{\smallskip \noindent \textbf{#1.}}

\usepackage[caption=false,font=footnotesize]{subfig}
\usepackage[hidelinks]{hyperref}
\ifCLASSINFOpdf
\else
\fi
\begin{document}
%\title{DECN: Automated Evolutionary Algorithms via Evolution Inspired Deep Convolution Network}
\title{DECN: Evolution Inspired Deep Convolution Network for Black-box Optimization}
%\title{Black-box Optimization with Few Budget}
% a separate \thanks must be used for each subsubsection as LaTeX2e's \thanks
% was not built to handle multiple subsubsections
%\author{Anonymous}
%\iffalse
\author{Kai~Wu,~\IEEEmembership{Member,~IEEE,}
        Xiaobin~Li
        Penghui~Liu,         and~Jing~Liu,~\IEEEmembership{Senior~Member,~IEEE}% <-this % stops a space
\IEEEcompsocitemizethanks{\IEEEcompsocthanksitem P. Liu and J. Liu are with Guangzhou Institute of Technology, Xidian University, Guangzhou 510555, China.
%\protect\\
E-mail: neouma@mail.xidian.edu.cn.
\IEEEcompsocthanksitem K. Wu is with School of Artificial Intelligence, Xidian University, Xi'an 710071, China.
E-mail: kwu@xidian.edu.cn.
%\IEEEcompsocthanksitem Corresponding author: Kai Wu.
}% <-this % stops a space

}
%\fi

\markboth{Journal of \LaTeX\ Class Files,~Vol.~14, No.~8, August~2015}%
{Shell \MakeLowercase{\textit{et al.}}: Bare Advanced Demo of IEEEtran.cls for IEEE Computer Society Journals}



\IEEEtitleabstractindextext{%
\begin{abstract}
Evolutionary algorithms (EAs) have emerged as a powerful framework for optimization, especially for black-box optimization. Existing evolutionary algorithms struggle to comprehend and effectively utilize task-specific information for adjusting their optimization strategies, leading to subpar performance on target tasks. Moreover, optimization strategies devised by experts tend to be highly biased. These challenges significantly impede the progress of the field of evolutionary computation. Therefore, 
%this paper introduces Automated EA: Automated EA leverages the inherent structure of the problem to automatically generate update rules (optimization strategies) for generating and selecting potential solutions, enabling the movement of a random population towards the optimal solution.
this paper first introduces the concept of Automated EA: Automated EA exploits structure in the problem of interest to automatically generate update rules (optimization strategies) for generating and selecting potential solutions so that it can move a random population near the optimal solution. However, current EAs cannot achieve this goal due to the poor representation of the optimization strategy and the weak interaction between the optimization strategy and the target task. We design a deep evolutionary convolution network (DECN) to realize the move from hand-designed EAs to automated EAs without manual interventions. DECN has high adaptability to the target task and can obtain better solutions with less computational cost. DECN is also able to effectively utilize the low-fidelity information of the target task to form an efficient optimization strategy.
The experiments on nine synthetics and two real-world cases show the advantages of learned optimization strategies over the state-of-the-art human-designed and meta-learning EA baselines. In addition, due to the tensorization of the operations, DECN is friendly to the acceleration provided by GPUs and runs 102 times faster than EA.
\end{abstract}

}

\maketitle

% is not selected <OR> if conference mode is selected - because compsoc
% conference papers position the abstract like regular (non-compsoc)
\IEEEdisplaynontitleabstractindextext

% \fi
\IEEEpeerreviewmaketitle

\ifCLASSOPTIONcompsoc
\IEEEraisesectionheading{\section{Introduction}\label{sec:introduction}}
\else
\section{Introduction}
\label{sec:introduction}
\fi


\IEEEPARstart{O}{ptimization} has a rich history and holds a fundamental place in various fields, including computer vision, machine learning, and natural language processing, where many tasks can be abstracted as optimization challenges. Among these tasks, black-box optimization, such as neural architecture search \cite{elsken2019neural} and hyperparameter optimization \cite{hutter2019automated}, is prevalent. Over time, numerous evolutionary algorithms (EAs) have been introduced to address these challenges. These EAs encompass genetic algorithms (GAs) \cite{mitchell1998introduction,Jin2029data,NEURIPS2018_85fc37b1,zhang2007moea,such2017deep,stanley2019designing} and evolution strategies (ES) \cite{wierstra2014natural,vicol2021unbiased,hansen2001completely,Auger2005restart,salimans2017evolution}. 

Recent research has showcased the advantages of meta-learning optimization strategies from data. For instance, Shala et al. \cite{shala2020learning} employed meta-learning to derive a policy for configuring mutation step-size parameters within CMA-ES \cite{hansen2001completely}. In a similar vein, Lange et al. \cite{lange2023discovering} introduced LES, which utilizes a self-attention-based search strategy to uncover effective update rules for two learning rates in diagonal Gaussian evolution strategies through CMA-ES \cite{hansen2001completely}. Notably, LES demonstrated superior performance compared to several ES baselines. Building upon this paradigm, subsequent work by Lange et al. \cite{lange2023arxiv} extended the framework to discover update rules for Gaussian genetic algorithms via Open-ES \cite{salimans2017evolution}. These studies collectively advocate for a promising approach to distilling versatile optimization strategies: initially amass a diverse dataset of tasks, then employ meta-learning to extract a policy from this data.

However, these methods exhibit two key limitations. Firstly, they are confined to meta-learning only a select set of hyperparameters for well-established evolutionary operators (such as Gaussian ES or GA), rather than comprehensively acquiring the overarching update rules necessary to generate and select potential solutions. Secondly, due to the inherently opaque nature of evaluating the performance of LES or LGA with predetermined parameters, the optimization problems designed to steer the development of optimization strategies tend to be costly and nonconvex, posing significant challenges, particularly in the context of high-dimensional problems. Furthermore, these approaches entail the sharing of weights among learned modules for each generation during the search process, which can be both rigid and inefficient \cite{tanabe2014improving,igel2003operator,Auger2005restart}. Consequently, current learned EAs suffer from suboptimal representation of optimization strategies, thereby hindering their ability to effectively harness task-specific information for the formulation of efficient optimization strategies.

We introduce the Deep Evolution Convolution Network (DECN) as a solution to address the aforementioned constraints. At the heart of any evolutionary algorithm lies the crucial task of producing and selecting potential solutions. In contrast to LES and LGA, which focus on learning the hyperparameters of existing solution-generation and solution-selection operations, we put forth a novel approach involving two tensor modules: a Convolution-Based Reasoning Module (CRM) and a Selection Module (SM). CRM facilitates the exchange of information among individuals within the population to generate potential solutions. We devise a grid-like framework that organizes the population into modified convolution operators and utilize mirror padding \cite{goodfellow2016deep} to generate these potential offspring. Meanwhile, SM updates the population by favoring the survival of the fittest solutions through pairwise comparisons between the offspring and the input population with respect to their fitness levels. This process is executed via the application of the mask operator.

Initially, we craft the Evolution Module (EM) by leveraging the functionalities of CRM and SM to emulate a single generation of Evolutionary Algorithms (EAs). To circumvent the necessity of employing meta-learning architectures that entail repeated execution of LGA or LES for parameter evaluation, we establish an end-to-end optimization framework by stacking multiple EMs, some with weight sharing and some without. Consequently, the training of DECN does not involve such complexities, effectively rendering it a neural network training problem. DECN possesses the inherent capacity to autonomously construct comprehensive update rules, thereby enhancing its representation capabilities.

The untrained DECN faces the challenge of optimizing the problem effectively since it lacks prior information about the target function. To address this, we devise a specific loss function geared towards maximizing the dissimilarity between the initial and output populations across various tasks, thereby facilitating the training of DECN towards optimal solutions. While traditional optimization approaches like reinforcement learning \cite{li2016learning} or evolution strategies \cite{lange2023arxiv} can directly optimize this loss function, they face limitations when applied to deep DECN due to its substantial parameter count. We have identified that the black-box nature of the loss function primarily stems from the inherent opacity of the task itself. To overcome this, we construct a differentiable and computationally efficient surrogate function set for the target black-box function. This approach circumvents the need for expensive function queries, enabling us to access valuable information about the target function and its gradient.

We evaluate the performance of DECN across a diverse set of scenarios, including nine synthetic functions, the mechanical arm motion planning problem, and neural network training. DECN consistently outperforms state-of-the-art (SOTA) evolutionary algorithms in all of these cases. Notably, DECN exhibits the ability to autonomously acquire efficient optimization strategies, both on high-fidelity and low-fidelity training function sets related to the target task. Moreover, DECN showcases its versatility by successfully transferring its learned capabilities to unseen objective functions with varying dimensions and populations, underscoring its robust generalization capacity. Finally, DECN's computational efficiency is highlighted, as it seamlessly harnesses Graphics Processing Unit (GPU) acceleration. When executed on a single 1080Ti GPU, DECN achieves a runtime that is 102 times faster than standard evolutionary algorithms.

\textbf{Contributions}. Current evolutionary algorithms struggle to effectively utilize task-specific information for optimizing strategies, resulting in subpar performance on target tasks. Additionally, strategies devised by experts often carry biases, further hindering progress in the field of evolutionary computation. Therefore, this paper introduces and achieves Automated EA: Automated EA leverages the inherent structure of the problem to automatically generate update rules (optimization strategies) for generating and selecting potential solutions, enabling the movement of a random population towards the optimal solution. The proposed DECN represents the first comprehensive framework of an automated evolutionary algorithm, bridging the gap from manually designed evolutionary algorithms to automated evolutionary algorithms, and holds paramount significance for the evolutionary computation field. The performance of DECN surpasses current state-of-the-art evolutionary algorithms and meta-learning EAs, showcasing rapid convergence to improved solutions.

The rest of this paper is organized as follows. Section 2 reviews human-designed genetic operators. formalizes the studied problem and introduces the background of DNMF. Section 3 presents the DECN framework. Section 4 conducts extensive experiments to evaluate the effectiveness of DECN. Finally, Section 5 concludes the whole paper.

\section{Reviews of Human-designed Genetic Operators}

\subsection{Problem Definition}
The optimization problem $f$ can be transformed or represented by a minimization problem, and constraints may exist for corresponding solutions:
\begin{equation}\label{eq:1}
\min \ f(s|\xi), s.t. \ x_i \in [d_i,u_i], \forall x_i \in s,
\end{equation}
where $s=(x_1, x_2, \cdots, x_D)$ represents the solution of $f$ while $d = (d_1, d_2,\cdots, d_D)$ and $u = (u_1, u_2, \cdots, u_D)$ denote the corresponding lower and upper bounds of the solution’s domain, respectively. $\xi$ is the known parameters of $f$. 
Suppose $n$ individuals of one population ($S=\{s_1, \cdots, s_n\}$) be $s_1=(x_1^1, x_2^1,\cdots, x_D^1), \cdots, s_n=(x_1^n, x_2^n, \cdots, x_D^n)$.

\subsection{Operators to generate offspring}
In EAs, there are many widely used crossover operators, such as one-point crossover operator, multi-point crossover operator, and linear crossover operator. The linear crossover operator is an extension over the one-point and multi-point crossover operators, usually conducted on $n$ individuals. Suppose $n$ individuals be $s_1=(x_1^1, x_2^1,\cdots, x_D^1), s_2=(x_1^2, x_2^2,\cdots, x_D^2),\cdots, s_n=(x_1^n, x_2^n,\cdots, x_D^n)$, then the linear crossover operator generates a new individual $s^{'}$ by the weight recombination as shown in Eq. \ref{eq:2}.
\begin{equation}\label{eq:2}
s^{'} = \sum_{i=1}^n {\lambda_is_i}, \ \ s.t. \ \sum_{i=1}^n {\lambda_i}=1
\end{equation}
Moreover, the subtraction is also applicable during the preproduction of a new individual, such as differential evolution (DE), usually conducted on $n$ individuals. We can reproduce a unique individual $s^*$ based on Eq. \ref{eq:3} with $s_1, s_2, \cdots, s_n$.
\begin{equation}\label{eq:3}
s^*=s_k + \sum_{i=2}^{n-1} F_i(s_i-s_{i+1})
\end{equation}
where $F_i$ is a scaling factor and $s_k$ is the best solution or is selected from ${s_1, s_2, \cdots, s_n}$. After an expression expansion, Eq. \ref{eq:2} and Eq. \ref{eq:3} can be summarized by a weighted recombination process as given in Eq. \ref{eq:4}. After that, a paradigm of crossover operators is obtained, and many available human-designed crossover operators follow an equivalent reproduction process. However, these human-designed crossover operators are usually designed with different parameters ($a_i$) based on the expert’s experience.
\begin{equation}\label{eq:4}
s^* = a_1 \times s_1 + a_2 \times s_2 + \cdots +a_n \times s_n = \sum_{i=1}^{n} {a_i \times s_i} 
\end{equation}

\subsection{Operators to select offspring}
Many different selection operators exist, such as the roulette-wheel selection operator in Eq. \ref{eq:5} and the binary tournament mating selection operator in Eq. \ref{eq:6}. The selection operator is to retain individuals with higher quality to the next generation, which can be regarded as an information selection process.
\begin{equation}\label{eq:5}
p_i = \frac{\varphi(f(s_i))}{\sum_i \varphi(f(s_i))}
\end{equation}
\begin{equation}\label{eq:6}
p_i = \left\{
\begin{matrix}
1 & f(s_i) < f(s_k)\\
0 & f(s_i) > f(s_k)
\end{matrix} \right.
, \ \ (s_i,s_k) \in S
\end{equation}
where $p_i$ reflects the probability that $s_i$ is selected for the next generation, $\varphi(f(s_i))$ in Eq. \ref{eq:5} transforms $s_i$’s fitness value $f(s_i)$ to be adapted to the roulette-wheel selection and $(s_i,s_k)$ in Eq. \ref{eq:6} are randomly selected from the population $S$. The selection process will be repeatedly conducted until the expected number of individuals are chosen.

\section{DECN}
\partitle{Notations} $\theta$ is the parameters of DECN. $S_0$ is the initial population, and $S_t$ is the output population of DECN. The goal of this paper is to enable an DECN with parameter $\theta$ to acquire a policy for target task using the information of other tasks.

\subsection{Training Dataset}
\textbf{Definition 1 Fidelity \cite{kandasamy2016gaussian}} \textit{Suppose the surrogate functions $f_1, f_2, \cdots, f_m$ are the continuous exact approximations of the black-box function $f$. We call these approximations fidelity, which satisfies the following conditions:
\begin{enumerate}
    \item $f_1, \cdots, f_i, \cdots, f_m$ approximate $f$. $||f-f_i||_{\infty} \leq \zeta_m$, where the fidelity bound $\zeta_1>\zeta_2> \cdots \zeta_m$.
    \item Estimating approximation $f_i$ is cheaper than estimating $f$. Suppose the query cost at fidelity is $\lambda_i$, and $\lambda_1<\lambda_2< \cdots \lambda_m$.
\end{enumerate}
}

We establish a function set $\mathcal{D}$ to train DECN. $\mathcal{D}$ only contains $(S_0, f_i(s|\xi))$, the initial population and objective/surrogate function, respectively.

\partitle{High-fidelity training dataset} We show the designed high-fidelity training functions as follows: $\mathcal{D} = \{f_m(s|\xi_{1}^{train}),\cdots, f_m(s|\xi_{i}^{train})\}$, 
where $f_{m}$ represents a set of high-fidelity functions related to the optimization objective $f$. $\xi_{i}^{train}$ represents the $i$th different values of $\xi$ in $f_m$, which is true for any index pair. The initial population $S_0$ is always randomly generated before optimization. 

\partitle{Low-fidelity training dataset} Similarly, the low-fidelity training dataset is as follows: $\mathcal{D} = \{f_1(s|\xi_{1}^{train}),\cdots, f_1(s|\xi_{i}^{train})\}$, where $f_{1}$ represents a set of low-fidelity functions.

The origins of surrogate functions are highly diverse, and we have not imposed any restrictions. They could encompass: 1) neural networks, Gaussian processes, regressions, etc., that approximate the target function; 2) existing standard function datasets, such as CEC 2013 \cite{liang2013problem} and COCO \cite{hansen2021coco}; 3) Other real-world scenarios.

\subsection{The Structure of DECN}
\begin{figure}[htbp]
%\vskip 0.2in
\begin{center}
\centerline{\includegraphics[width=\linewidth]{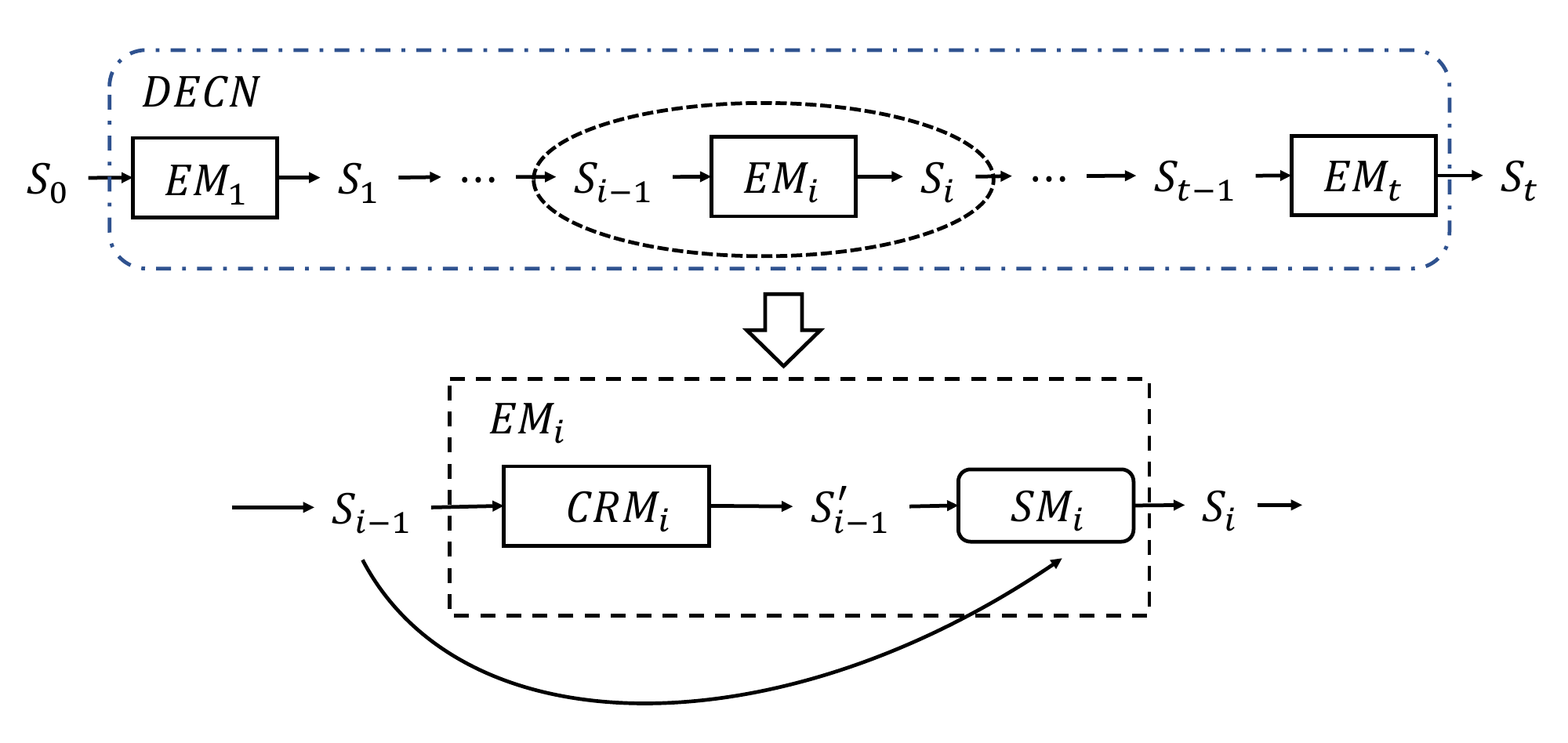}}
\caption{A general view of DECN and EM.}
\label{fig:10}
\end{center}
\vskip -0.3in
\end{figure}

In Figure \ref{fig:10}, a EM based on CRM and SM is designed to learn optimization strategies. Then, DECN is established by stacking several EMs. $S_{i-1}$ is the input population of $EM_i$. $S_{i-1}^{'}$ is the output of CRM in order to further improve the quality of individuals in the global and local search scopes. Then, SM selects the valuable individuals from $S_{i-1}$ and $S_{i-1}^{'}$ according to their function fitness.

\subsubsection{Convolution-based Reasoning Module}

We design CRM to ensure that individuals in the population can exchange information to generate the potential solutions near the optimal solution. The corresponding correction to the convolution operator can achieve this goal. 
%The capability to reason over available information is essential for learning to optimize. This part mainly describes (1) \emph{the shared commonalities between convolution operator and mutation operators in DE}; (2) \emph{how to construct CRM to generate new offspring}.

\partitle{Motivation} CRM can be likened to the parameterized mutation operator in differential evolution, which is utilized to achieve the function of producing potential offspring. For a population of size $n^2$ with a 1D dimension, we arrange individuals in an $n \times n$ matrix. In this case, Eq. \ref{eq:4} can be transformed into $s^* = \sum_{i=1,j=1}^{n} {a_{i,j} \times s_{i,j}}$, which resembles a convolution operator. The formulation of the convolution operator is given as follows,
$$g(i,j)= \sum_{k,l}f(k,l)h(i+k,j+l)$$
$$s.t. \ \ {f(k,l)=w_{k,l}, \ h(i+k,j+l)=p_{i+k,j+l}}$$
where $p_{i,j}$ denotes the pixel value in the input image if we handle the image task, $(i, j)$ locates the position of the pixel, $g(i,j)$ denotes the output values (feature) of convolution, and $w_{k,l}$ represents the related parameters of convolution kernels.

%\partitle{Organize Population into Convolution}
We arrange all individuals in a lattice-like environment with a size of $L\times L$. In this case, we represent the population by using a tensor $(i, j, d)$, where $(i, j)$ locates the position of one individual $S(i,j)$ in the $L \times L$ lattice and $d$ is the dimension information of this individual. The individuals in the lattice are sorted in descending order to construct a population tensor with a consistent pattern. The number of channels in input tensors is $D+1$, where $D$ is the dimension of the optimization task, and the fitness of individuals occupies one channel. The fitness channel does not participate in the convolution process but is essential for the information selection in the selection module.

The traditional convolution operator usually merges the information simultaneously among different channels and locations. Although a population can be organized as a tensor, the standard convolution operator cannot merge information in different dimensions similar to recombination operators.
After organizing the population into a tensor $(L, L, D+1)$, we modified the depthwise separable convolution (DSC) operator \cite{chollet2017xception} to generate new individuals by merging information in different dimensions among individuals. The DSC operator includes a depthwise convolution followed by a pointwise convolution. 
%We employ the DSC operator to merge information in different dimensions similar to recombination operators.
%The DSC operator is a widely employed convolution method to reduce computational costs. The DSC operator recombines the neighborhood information while separating the information into different dimensions. If convolving each channel, it is the depthwise convolution. 
Pointwise convolution maps the output channel of depthwise convolution to a new channel space. When applied to our task, we remove the pointwise convolution in DSC to avoid the information interaction between channels.
Eq. (\ref{eq:8}) provides the details about how to reproduce offspring based on parent $S$.%, and one example is shown in Appendix Figure \ref{fig:a3}.
\begin{equation}\label{eq:8}
S^{'}(i,j)= \sum_{k,l}{w_{k,l}S(i+k,j+l)},
\end{equation}
where $w_{k,l}$ represents the related parameters of convolution kernels. 
%An illustration of the DSC operator is provided in Figure \ref{fig:5}.
Moreover, to adapt to optimization tasks with different dimensions, different channels share the same parameters. The parameters within convolution kernels record the strategies learned by this module to reason over available populations given different tasks. There are still two critical issues to address here.

1) \textit{Since there does not exist a consistent pattern in the population, the gradient upon parameters is unstable as well as divergent.} A fitness-sensitive convolution is designed, where the CRM’s attention to available information should be relative to the quality and diversity of the population. $w_{k,l}$ reflects the module’s attention during reasoning and is usually relative to the fitness of individuals. After that, this problem is resolved by simply sorting the population in the lattice based on individuals’ fitness.

%DECN may achieve reasoning functions like recombination operators to reproduce individuals with higher quality in the global search range. More importantly, we can train the parameters within convolution kernels without worrying about unstable and divergent gradients.

2) \textit{The scale of the offspring.} We conduct mirror padding before the convolution operator to maintain the same scale as the input population. Mirror padding copies the individuals to maintain the same scale between the offspring and the input population. 
As the recombination process conducts the information interaction among individuals, copying the individual is better than extending tensors with a constant value. %An implementation of mirror padding to the population is given in Figure \ref{fig:a4} (see Appendix \ref{a2}). 

The size of convolution kernels within CRM determines the number of individuals employed to implement reasoning of $S^{'}(i,j)$. Several essential issues are necessary to be considered. After that, this paper employs convolution kernels with commonly used sizes. Different convolution kernels produce corresponding output tensors, while the final offspring are obtained by averaging multiple convolutions' output. Then, the fitness of this last offspring will be evaluated.
%For an individual $s_i$, its fitness value is evaluated by $f(cx(s_i))$, $cx$ denotes the convolution matrix. Suppose the domain of $x_j^i (j=1, \cdots, D)$ be $[d_j, u_j]$, if $x_j^i > u_j$, $x_j^i=u_j$; if $x_j^i < d_j$, $x_j^i=d_j$.

\partitle{Discussion}
1) \textbf{How many individuals should participate in the CRM reasoning progress}. It remains a challenge to implement information reasoning over multi-individuals in EAs. In most recombination operators, the participant number is usually set to 2. However, based on the gradient information provided by the back-propagation, it is easy to control an individual’s element by adjusting $w_{k,l}$.

2) \textbf{How to integrate the offspring produced by different convolution kernels}. Since the convolution operation can be transformed as a multiplication between matrices, simply averaging over the results output by different convolution kernels does not influence the training process. For example, $a_1C^1x_1^i + a_2C^2x_2^i + a_3C^3x_3^i \leftrightarrow  C^{'1}x_1^i + C^{'2}x_2^i + C^{'3}x_3^i$,
%Eq. \ref{eq:9}.
%\begin{equation}\label{eq:9}
%a_1C^1x_1 + a_2C^2x_2 + a_3C^3x_3 \leftrightarrow  C^{'1}x_1 + C^{'2}x_2 + %C^{'3}x_3
%\end{equation}
where $x_1^i$, $x_2^i$, and $x_3^i$ are input elements of $s_i$, $a_1$, $a_2$, and $a_3$ are the constant, and $C$ denotes the convolution matrix.

3) \textbf{How many convolution kernels should be used within CRM}. We suppose that these are three convolution kernels for $x$. We can find that the outcome $a_1C_{3\times3}^1x + a_2C_{3\times3}^2x + a_3C_{3\times3}^3x$ is equivalent to $a^{'}C_{3\times3}^{'}x$. The output of multiple convolution kernels can be replaced by one convolution kernel. Thus, the number of convolution kernels of the same size has no apparent influence on DECN.

%Based on $a_1C_{3\times3}^1x + a_2C_{3\times3}^2x + a_3C_{3\times3}^3x \leftrightarrow  a^{'}C_{3\times3}^{'}x$, we can find that the number of convolution kernels with the same size has no apparent influence upon DECN.

4) \textbf{The impact of neighborhood recombination operation}. The neighborhood recombination operation has been commonly accepted in EAs to alleviate the selection pressure and prevent the premature convergence of populations. Moreover, the receptive field of convolution kernels expands as the number of layers increases. Thus, DECN can learn efficient optimization strategies across generations.

\subsubsection{Selection Module}
SM updates individuals based on a pairwise comparison between the offspring and input population regarding their fitness for efficiency and simplicity. Thereafter, a matrix subtraction of fitness channel corresponding to $S_{i-1}$ and $S_{i-1}^{'}$ compares the quality of individuals from $S_{i-1}$ and $S_{i-1}^{'}$ pairwise. A binary mask matrix indicating the selected individual can be obtained based on the indicator function $l_{x>0}(x)$, where $l_{x>0}(x)=1$ if $x>0$ and $l_{x>0}(x)=0$ if $x<0$. 
%To extract selected individuals from $S_{i-1}$ and $S_{i-1}^{'}$, we construct a binary mask tensor by copying and extending the mask matrix to the same shape as $S_{i-1}$ and $S_{i-1}^{'}$.
The selected information forms a new tensor $S_i$ by employing Eq. (\ref{eq:12}).
\begin{equation}\label{eq:12}
\begin{split}
S_i &= tile(l_{x>0}(M_{F'}-M_F)) \bullet S_{i-1} \\
&+ tile(1-l_{x>0}(M_{F'}-M_F)) \bullet S_{i-1}^{'}
\end{split}
\end{equation}
where the \emph{tile} copy function extends the indication matrix to a tensor with size $(L, L, D)$, $M_F (M_{F'})$ denotes the fitness matrix of $S_{i-1} (S_{i-1}^{'})$, and $\bullet$ indicates the pairwise multiplication between inputs. 

\subsection{Training of DECN}
%DECN with $t$ EMs generates the offspring $S_t$ from the input population $S_0$ based on $S_t=G_{\theta}(S_0,f(s|\xi))$ and can be trained based on end-to-end mode. Then, given a proper loss function and training dataset, DECN can be trained to learn optimization strategies towards the objective function $f(s|\xi)$ by the back-propagation. 
\begin{algorithm}[htbp]
   \caption{Training of DECN}
   \label{alg:1}
\begin{algorithmic}
   \State {\bfseries Input:} Batch size for Adam, $\Omega$; Training dataset, $\mathcal{D}$;
   \State {\bfseries Output:} Parameters of DECN $\theta$;
   \State Randomly initialize $\theta$ of DECN;
   %\STATE Randomly initialize $\xi_{i}^{train}$ to adjust $f_j$ in ${F}^{train}$;
   \Repeat
   \State Randomly initialize a minibatch $\Omega$ comprised of $K$ populations $S_0$;
   \For{$f_i$ {\bfseries in} $\mathcal{D}$}
   \State Update $\theta$ by $\mathcal{L}_i$ given training data $(S_0, f_i)$;
   \EndFor
   \State Update $\theta$ by minimizing $-{1}/{m} \sum_i \mathcal{L}_i$;
   \State Sample new parameters $\xi_i$ of $f_i$ in $\mathcal{D}$ every $T$ epochs;
   \Until{training is finished}
\end{algorithmic}
\end{algorithm}

DECN attempts to search for individuals with high quality based on the available information. The loss function tells how to adaptively adjust the DECN parameters to generate individuals closer to the optimal solution. According to the Adam \cite{kingma2014adam} method, a minibatch $\Omega$ is sampled each epoch for the training of DECN upon $\mathcal{D}$, which is comprised by employing $K$ initialized $S_0$ for each $f_i$. We give the corresponding mean loss $\mathcal{L}_i$ of minibatch $\Omega$ for $f_i$ in $\mathcal{D}$,
\begin{equation}\label{eq:14}
\underset{\theta}{\arg\min} \ - \sum_{S_0 \in \Omega} \frac{ \frac{1}{|S_0|} {\sum \limits_{s\in S_0} {f_i(s|\xi)}} -\frac{1}{|G_{\theta}(S_0)|} {\sum \limits_{s \in G_{\theta}(S_0)}} f_i(s|\xi)}{\left|\frac{1}{|S_0|} {\sum \limits_{s\in S_0} f_i(s|\xi)}\right|}
\end{equation}
Eq. (\ref{eq:14}) is to maximize the difference between the initial population and the output population of DECN to ensure that the initial population is close to the optimal solution. Moreover, Eq. (\ref{eq:14}) are generally differentiable based on the constructed training dataset. The process of training DECN is showed in Algorithm 1.

Suppose the objective functions are hard to be formulated, and the derivatives of these functions are not available at training time, then two strategies can be employed: 1) Approximate these derivatives via REINFORCE \cite{williams1992simple}; 2) Use the ES \cite{vicol2021unbiased} method to train DECN.
Algorithm 1 provides the training process of DECN. After DECN has been trained, DECN can be used to solve the black-box optimization problems since the gradient is unnecessary during the test process.

\subsection{Underlying Principle of DECN}
DECN can be likened to differential evolution algorithms, wherein CRM functions as a parameterized mutation operator, and SM operates analogously to a roulette wheel selection operator. The utilization of the proposed loss function enables the adjustment of DECN's parameters to maximize its performance on the target function. In this context, DECN possesses the optimal strategy for optimizing the target task. If the target function is unobservable or costly, only proxies in the form of high-fidelity or low-fidelity approximations can be obtained. In such cases, we can fine-tune DECN's parameters to maximize its performance on the proxy function. Hence, DECN can achieve enhanced performance, which is a natural progression.

\begin{figure*}[htbp]
\centering
\subfloat[F4(Sphere), $D$=10] {\includegraphics[width=0.33\linewidth]{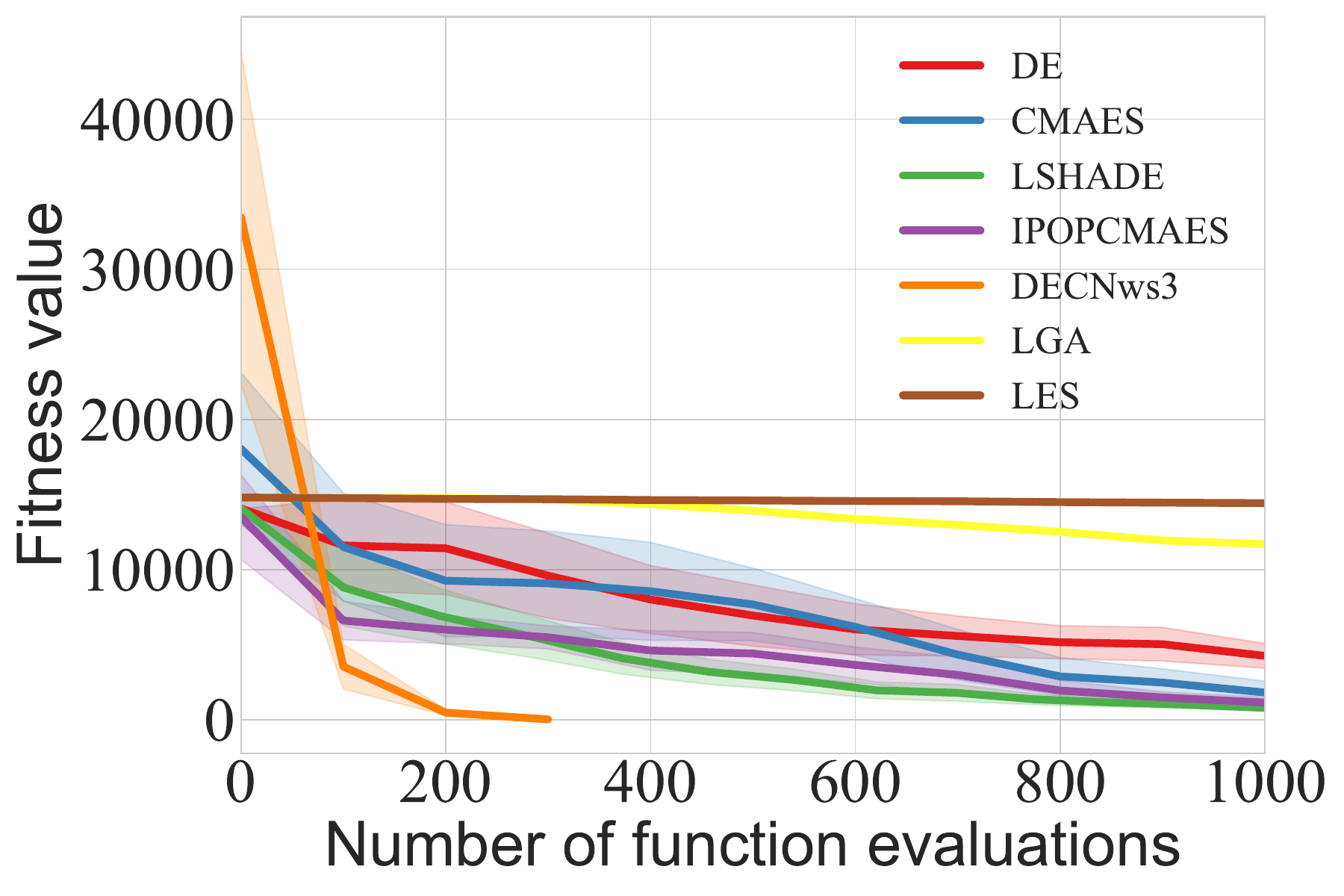}}
%	\hspace{-10pt}
\subfloat[F5(Mix), $D$=10] {\includegraphics[width=0.33\linewidth]{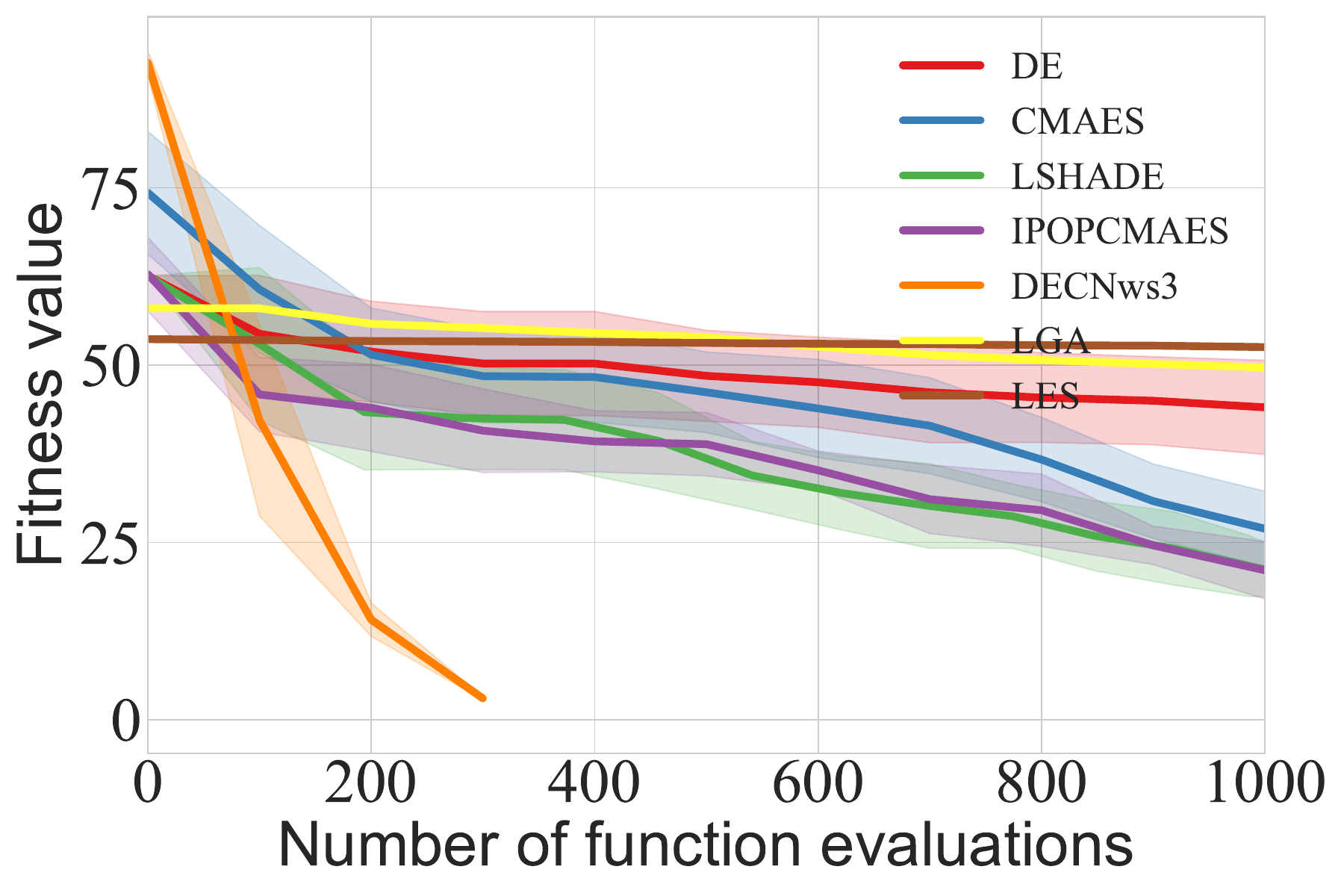}}
\subfloat[F6(Rosenbrock), $D$=10] {\includegraphics[width=0.33\linewidth]{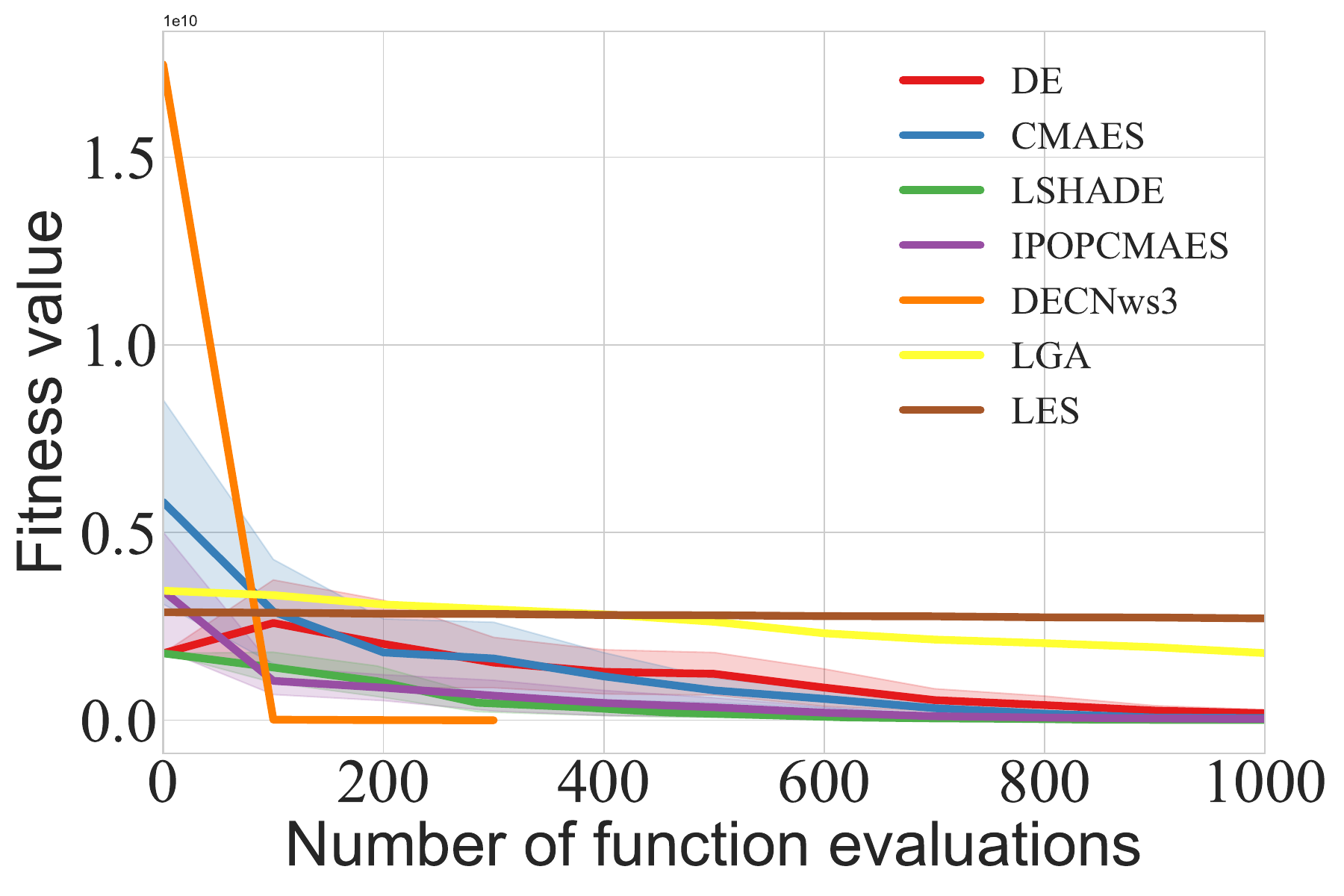}}\\
\subfloat[F7(Rastrigin), $D$=10] {\includegraphics[width=0.33\linewidth]{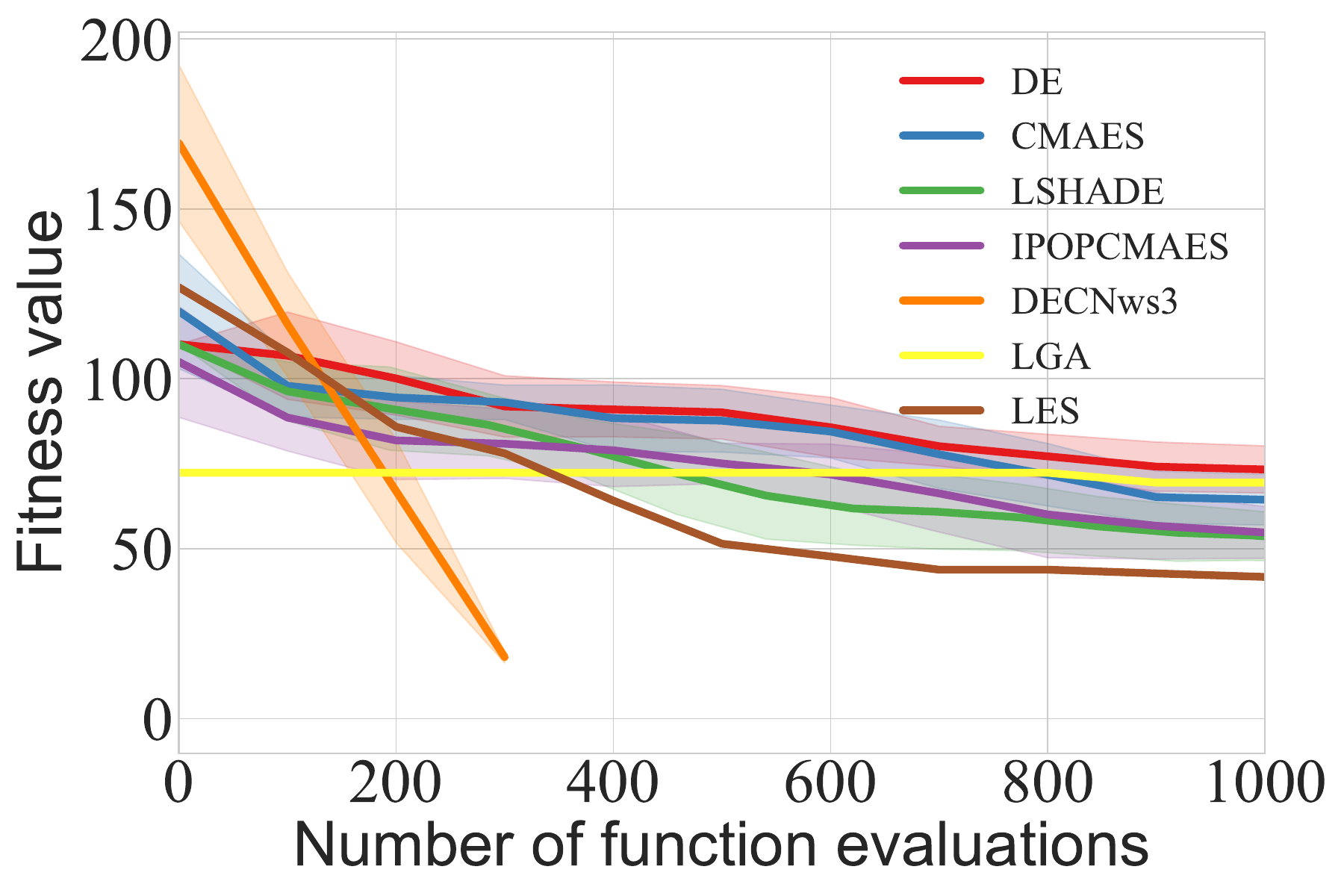}}
%	\hspace{-10pt}
\subfloat[F8(Griewank), $D$=10] {\includegraphics[width=0.33\linewidth]{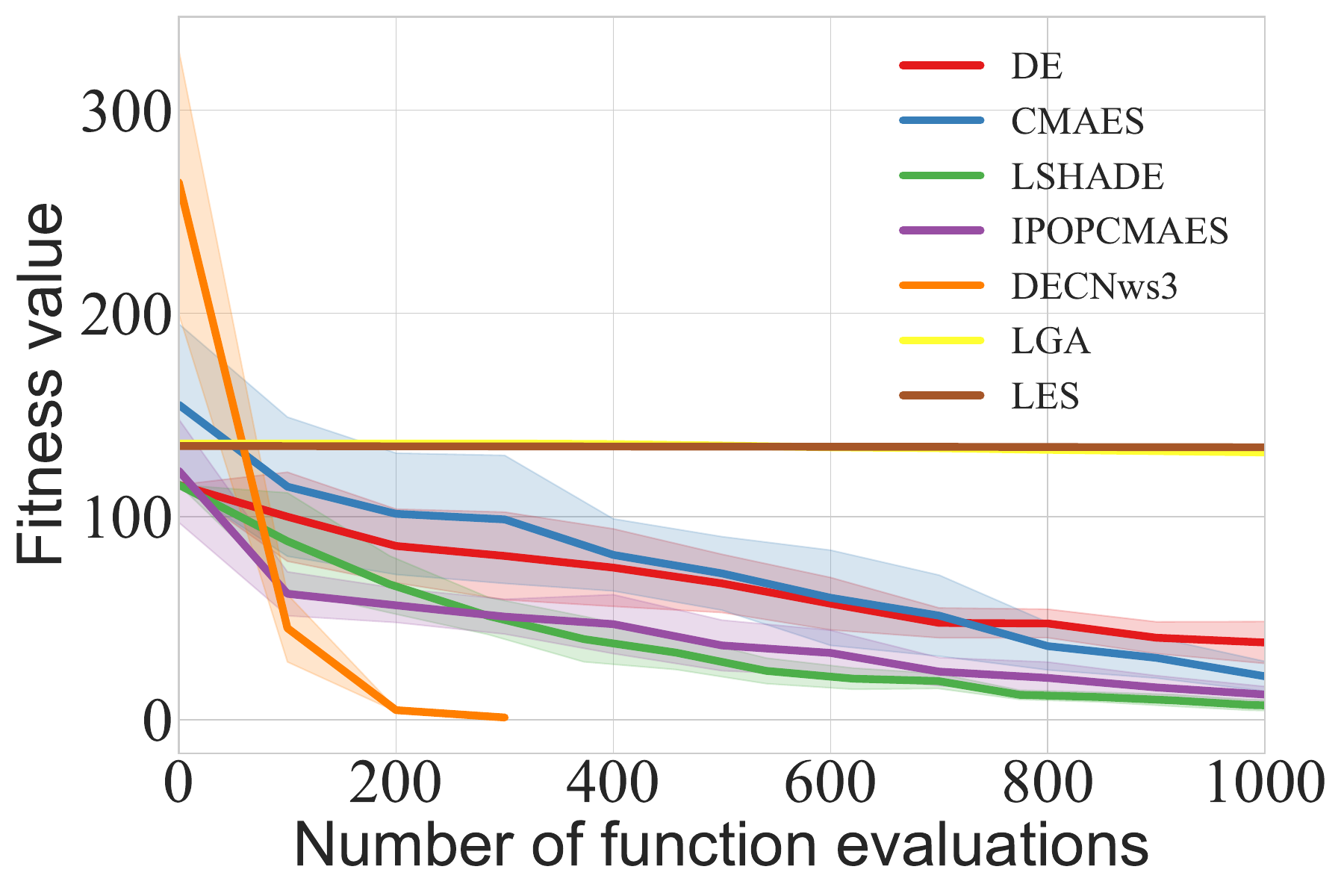}}
\subfloat[F9(Ackley), $D$=10] {\includegraphics[width=0.33\linewidth]{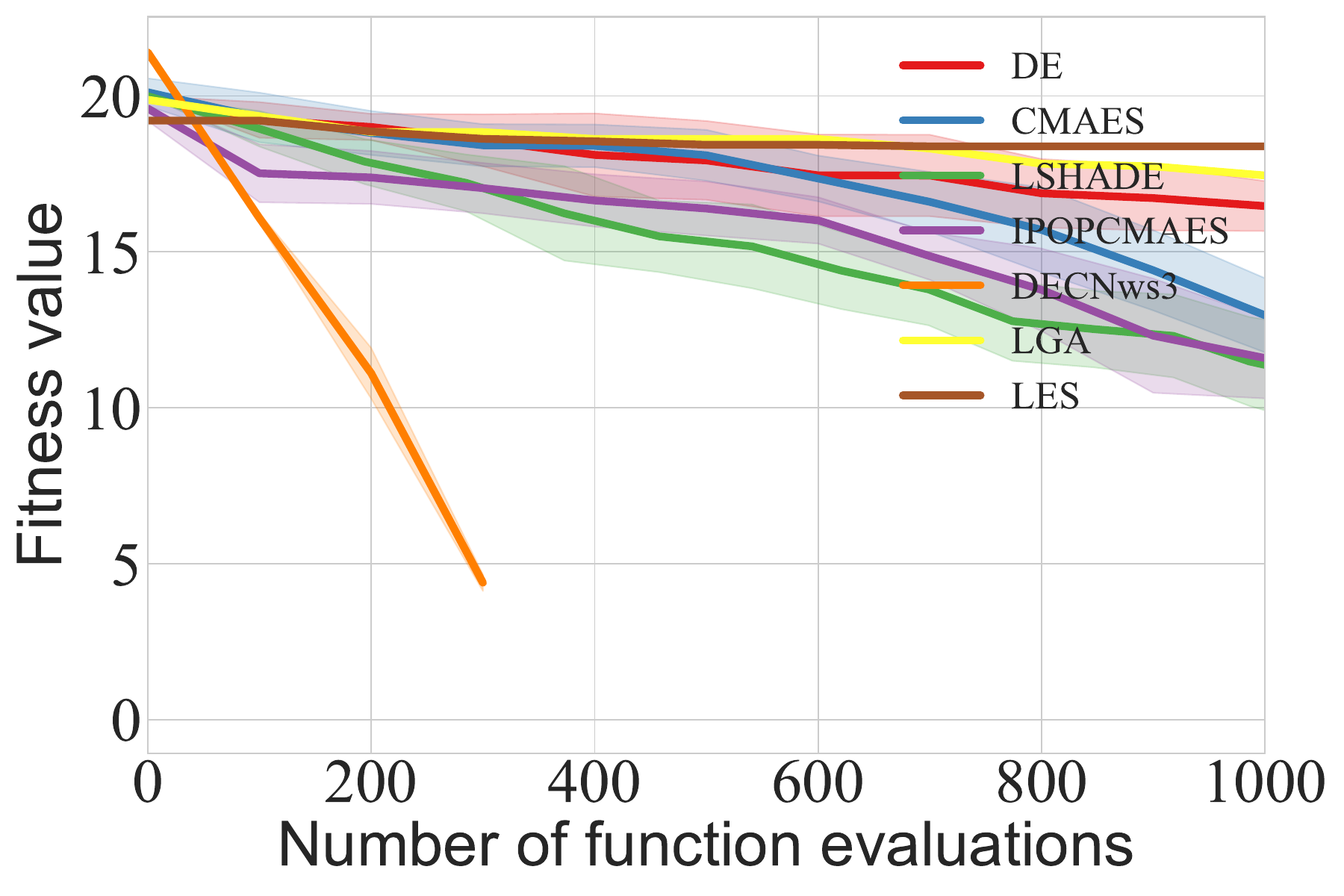}}\\
\caption{The compared results on high-fidelity surrogate functions. (a)-(f) represent the convergence curves of DECN and the comparison algorithms on F4-F9 when $D$=10.}
\label{fig:r1}
\end{figure*}

\begin{figure*}[ht]
\centering
\subfloat[F4(Sphere), $D$=100] {\includegraphics[width=0.33\linewidth]{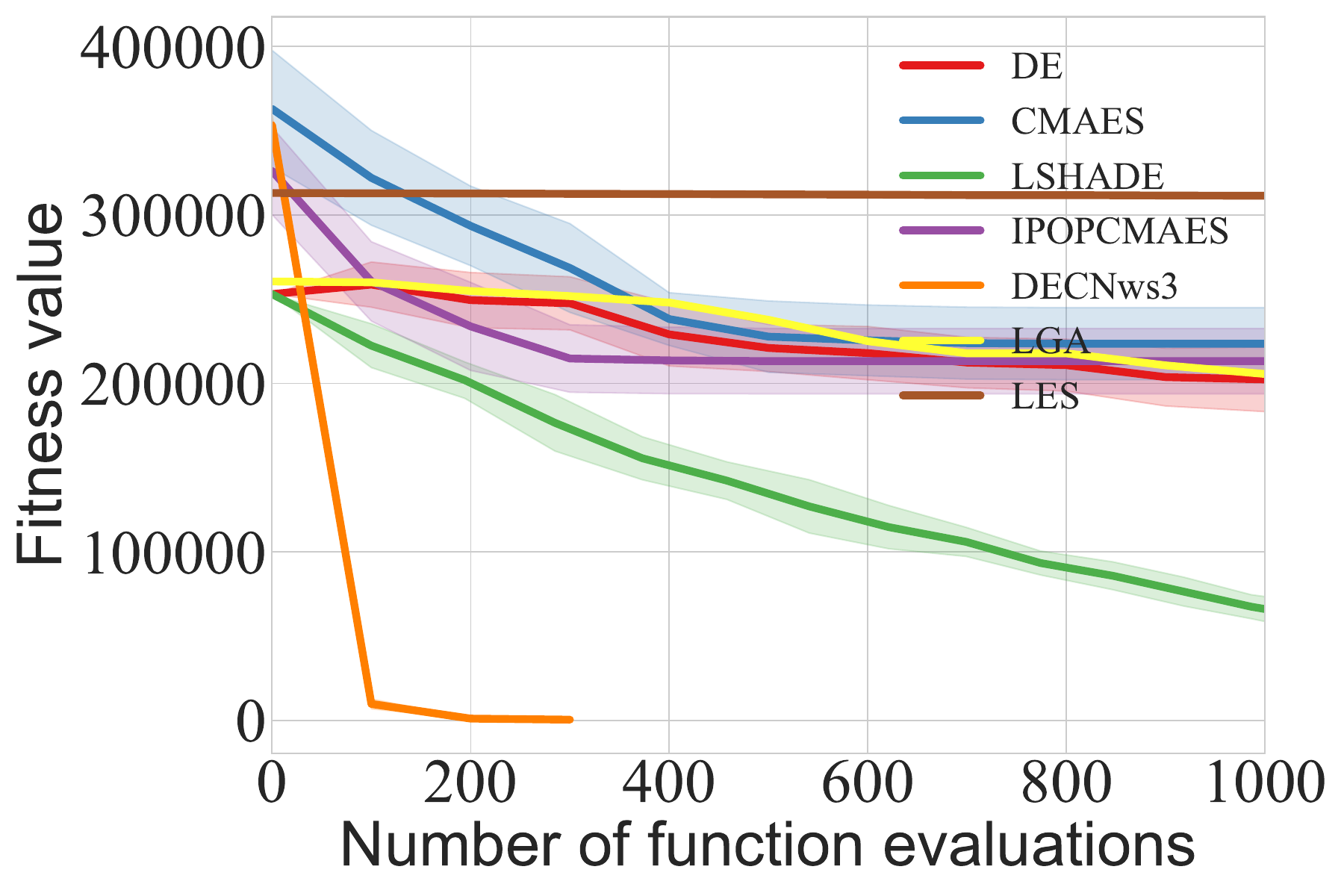}}
\subfloat[F5(Mix), $D$=100] {\includegraphics[width=0.33\linewidth]{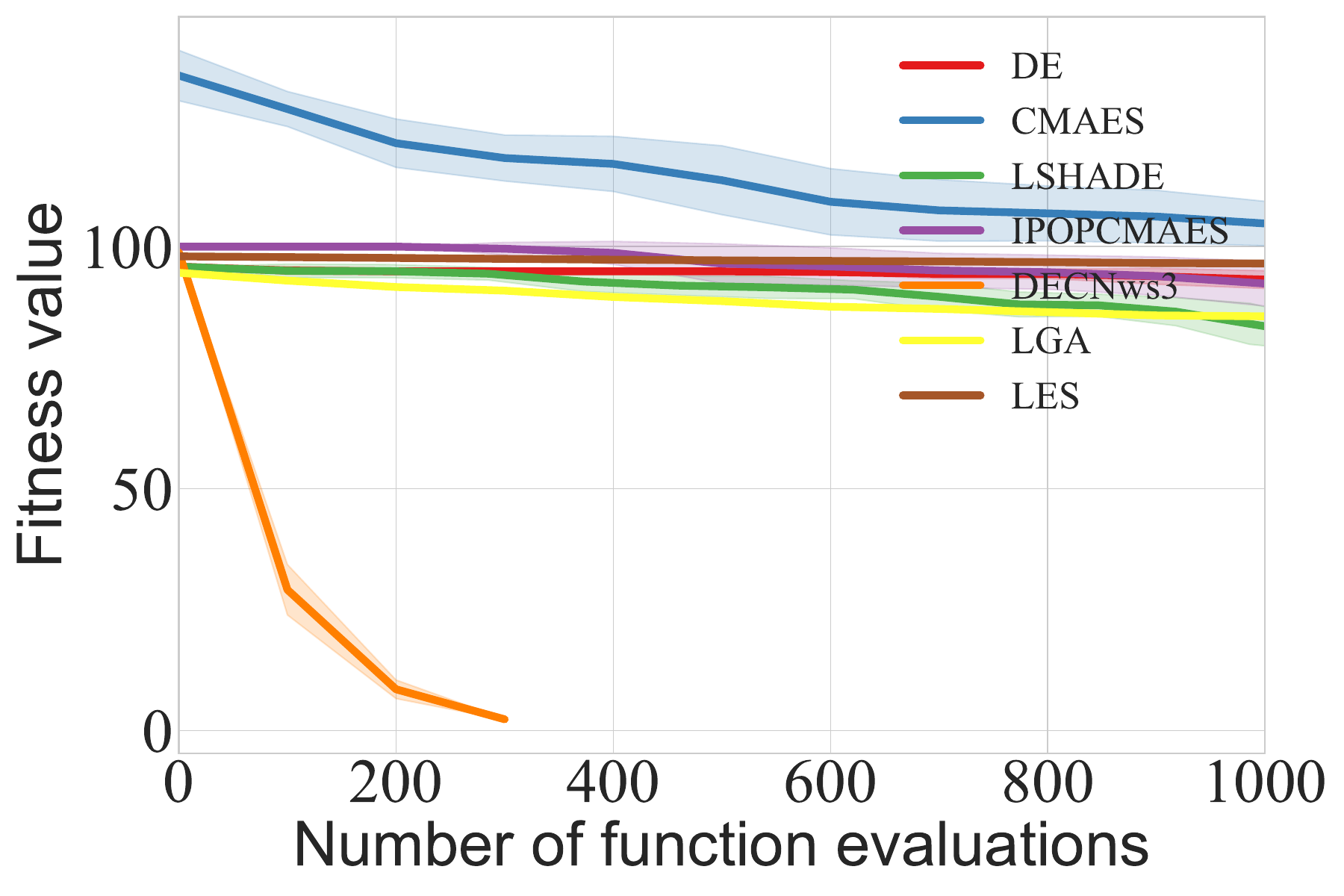}}
\subfloat[F6(Rosenbrock), $D$=100] {\includegraphics[width=0.33\linewidth]{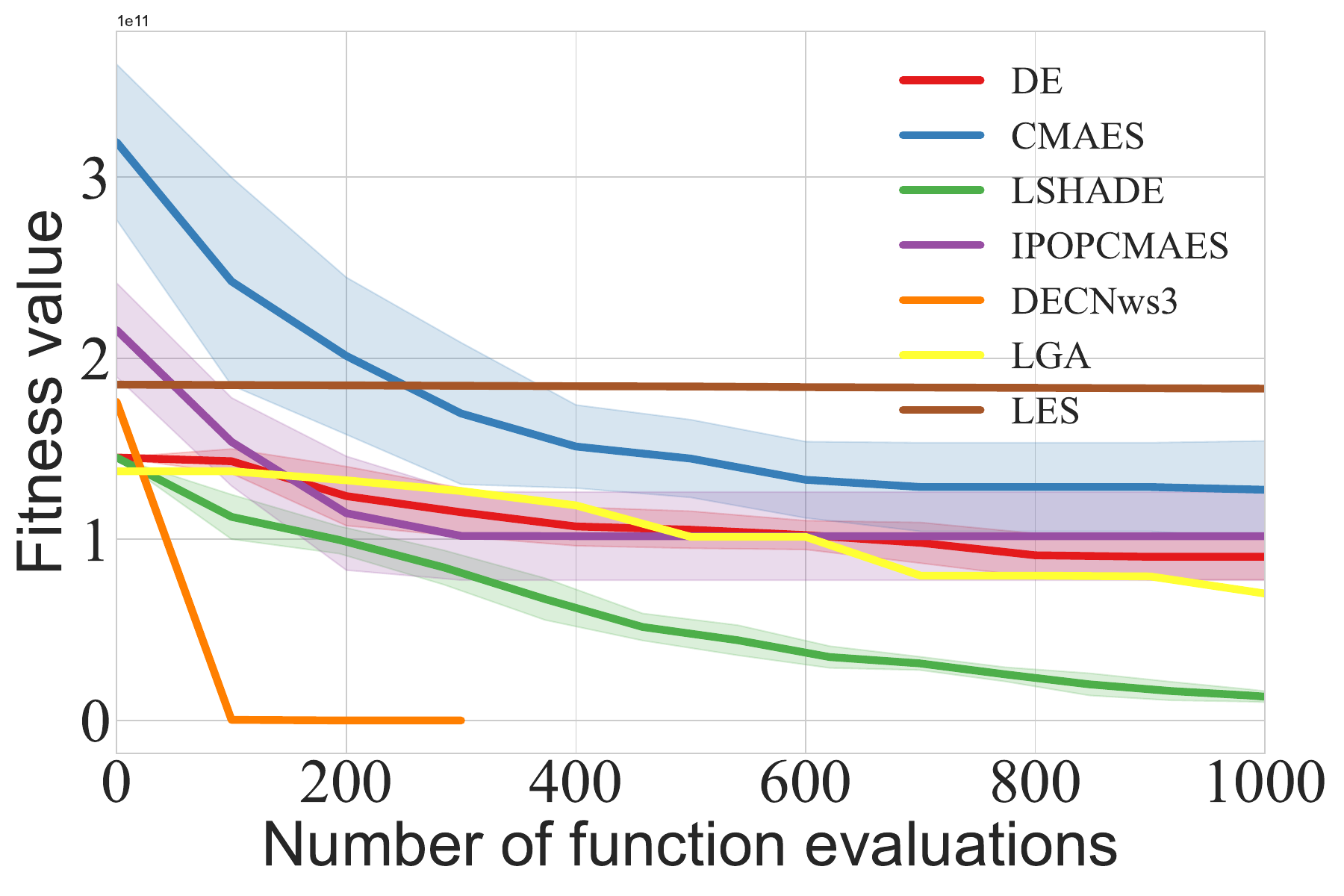}}\\
\subfloat[F7(Rastrigin), $D$=100] {\includegraphics[width=0.33\linewidth]{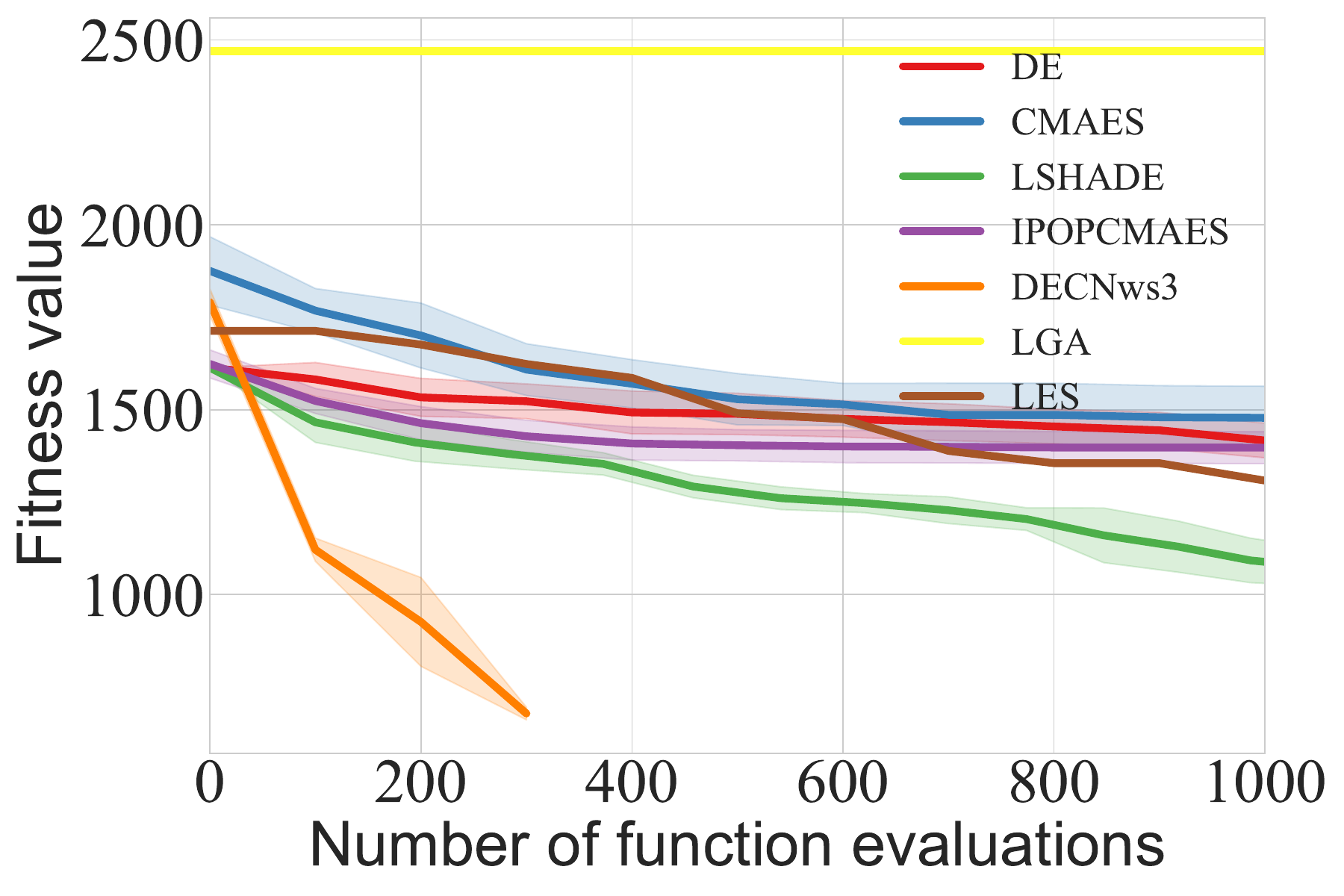}}
\subfloat[F8(Griewank), $D$=100] {\includegraphics[width=0.33\linewidth]{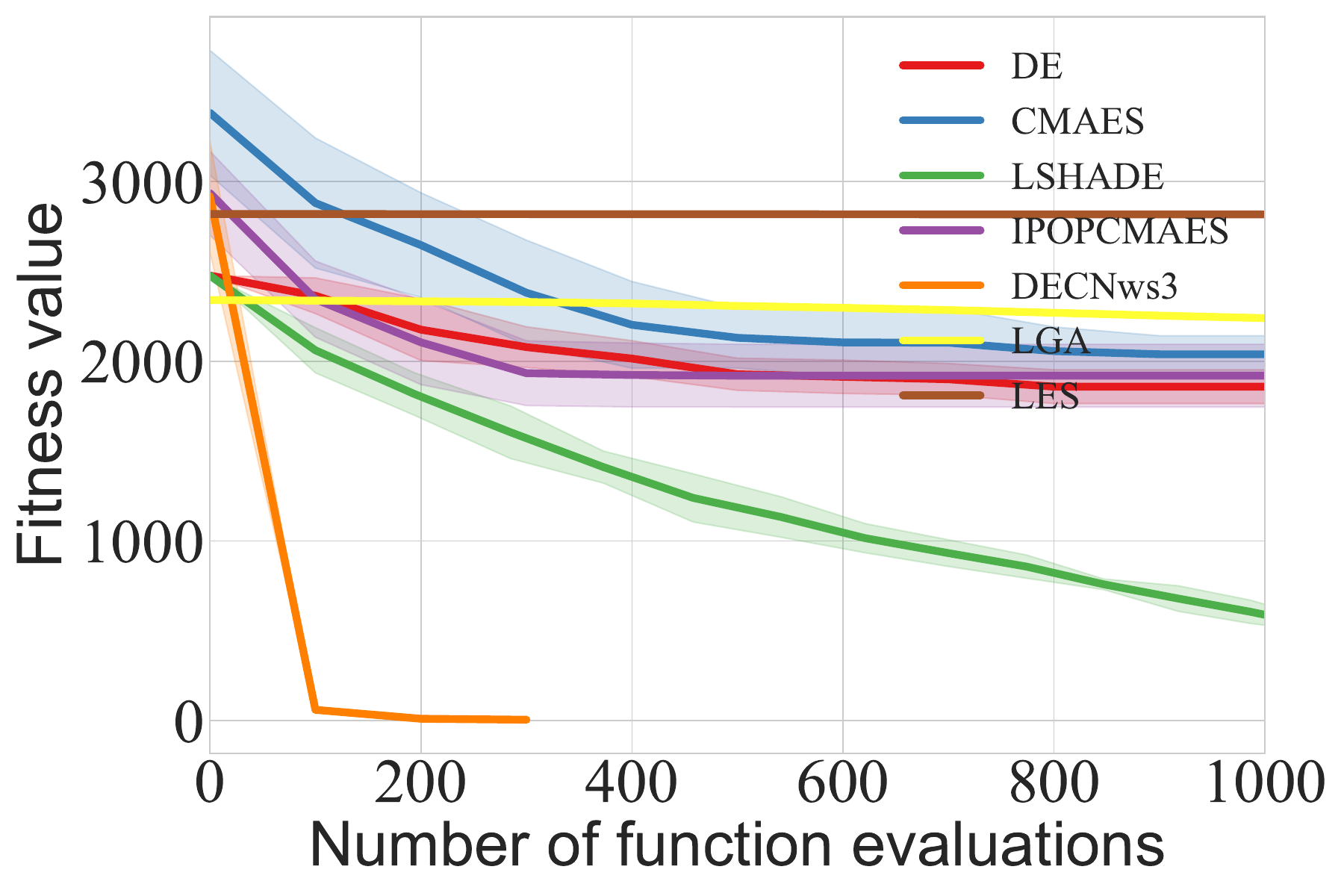}}
\subfloat[F9(Ackley), $D$=100] {\includegraphics[width=0.33\linewidth]{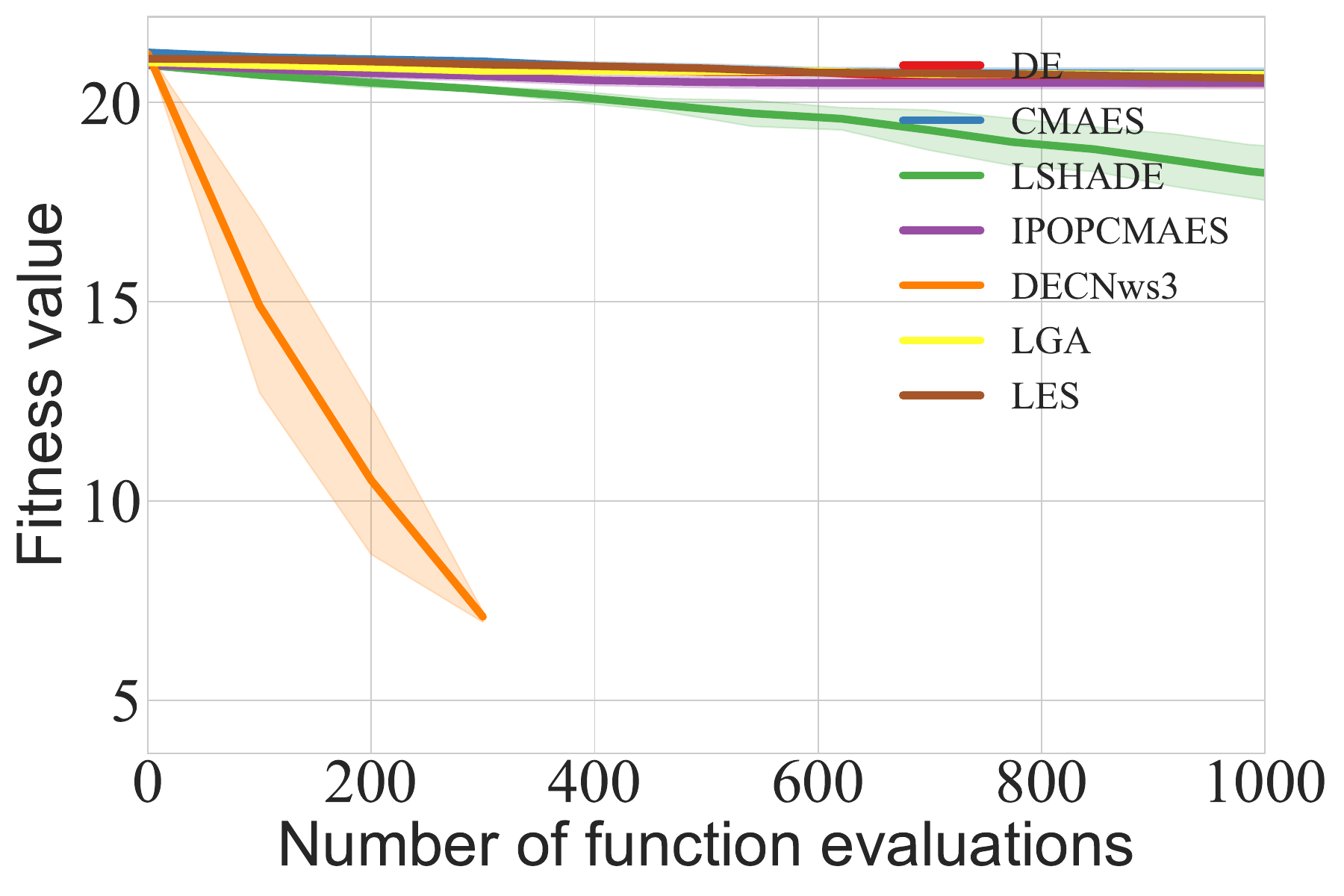}}\\
\caption{The compared results on high-fidelity surrogate functions. (a)-(f) represent the convergence curves of DECN and the comparison algorithms on F4-F9 when $D$=100.}
%\vskip -0.2in
\label{fig:r2}
\end{figure*}

\begin{table}[htbp]
\caption{Training functions. For all functions, $x \in [-10,10], b\in [-10,10]$.}
\label{table:a1}
\begin{center}
\begin{tabular}{lc}
\toprule
ID & Functions \\
\midrule
F1    & $\sum_i {|w_i sin(x_i-b_i)|}$ \\
F2    & $\sum_i {|x_i-b_i|}$ \\
F3    & $\sum_i {|(x_i-b_i)-(x_{i+1}-b_{i+1})|} + \sum_i {|x_i-b_i|}$ \\
\bottomrule
\end{tabular}
\end{center}
\end{table}

\begin{table*}[htbp]
\caption{Testing Functions. F4(Sphere); F5(Mix); F6(Rosenbrock); F7(Rastrigin); F8(Griewank); 9(Ackley). Here, $z_i=x_i-b_i$.}
\label{table:a2}
\scriptsize
\begin{center}
\begin{tabular}{lcc}
\toprule
ID & Functions & Range \\
\midrule
F4    & $\sum_i z_i^2$ & $x \in [-100,100], b\in [-50,50]$ \\
F5    & $\max\{|z_i|, 1 \le i \le D\}$ & $x \in [-100,100], b\in [-50,50]$ \\
F6    & $\sum \limits_{i=1}^{D-1} (100(z_i^2-z_{i+1})^2+{(z_i-1)}^2)$ & $x \in [-100,100], b\in [-50,50]$ \\
F7   & $\sum \limits_{i=1}^{D} (z_i^2-10\cos(2\pi z_i)+10),$ & $x \in [-5,5], b\in [-2.5,2.5]$ \\
F8    & $\sum \limits_{i=1}^{D} {\frac{z_i^2}{4000}}- \prod_{i=1}^D \cos(\frac{z_i}{\sqrt{i}})+1$ & $x \in [-600,600], b\in [-300,300]$ \\
F9    & $-20\exp(-0.2\sqrt{\frac{1}{D} \sum_{i=1}^D z_i^2})-\exp(\frac{1}{D} \sum_{i=1}^D \cos(2\pi z_i))+20+\exp(1)$ & $x \in [-32,32], b\in [-16,16]$ \\
\bottomrule
\end{tabular}
\end{center}
\end{table*}

\section{Experiments}

\subsection{Baselines}
\partitle{SOTA EA baselines} First, LSHADE \cite{tanabe2014improving} and I-POP-CMA-ES \cite{Auger2005restart}, two SOTA human-designed ES and differential evolution (DE) in most CEC and GECCO optimization competitions, are provided as the reference. Moreover, DE (DE/rand/1/bin) \cite{das2010differential} and CMA-ES are employed as baselines. These two are currently the best basic EAs. 

\partitle{Learned EA baselines} LES \cite{lange2023discovering} and LGA \cite{lange2023arxiv} are the SOTA learned EA baselines, which are employed to demonstrate the strong optimization strategy representation of DECN. The model parameters of LGA and LES are copyed from evosax \cite{evosax2022github} as suggested by the original paper, treating them as benchmark SOTA EAs.

\partitle{DECN} Here, we design three models, DECNws3, DECNws30, and DECNnws15. For example, DECNws3 contains 3 EMs, and the parameters of these EMs are consistent (weight sharing). DECNn15 does not share parameters across 15 EMs. The detailed parameters of these models and the setting of the baselines can be found in Appendix. Note that we explore whether DECN can obtain better solutions with less function evaluation cost. Therefore, the number of evaluations of all algorithms is set to be small. 

\begin{table*}[htb]
\caption{Experimental setup for DECNws30, DECNws3, and DECNnws15. In DECNws3, parameters of these three convolution kernels are consistent across different EMs (weight sharing). Moreover, during the training process, the 2-norm of gradients is clipped to be not larger than 10, and the learning rate ($lr=0.01$) shrinks every 100 epochs. The shrinking rate is set to 0.9.
The generation of these reference algorithms is set to 100, while DECNws3 only evolves the population with 3 EMs. 5000 epochs are conducted during the training process. All experimental studies are performed on a Linux PC with Intel Core i7-10700K CPU at 3.80GHz and 32GB RAM.}
\label{table:a3}
\begin{center}
    \begin{tabular}{cccccccccp{1cm}p{1cm}}
    \toprule
    Model & $L$ & $D$ & $K$ & EMs & Convolution kernels & $lr$ & Epochs & $T$ & Weight share & Gradient norm \\
    \multirow{3}[0]{*}{DECNws30} & \multirow{3}[0]{*}{10} & \multirow{3}[0]{*}{10} & \multirow{3}[0]{*}{32} & \multirow{3}[0]{*}{30} & $3\times 3$: $u=0,\sigma=0.5$ & \multirow{3}[0]{*}{0.01} & \multirow{3}[0]{*}{10000} & \multirow{3}[0]{*}{10} & \multirow{3}[0]{*}{True} & \multirow{3}[0]{*}{10} \\
          &       &       &       &       & $5\times 5$: $u=0,\sigma=0.5$ &       &       &       &       &        \\
          &       &       &       &       & $7\times 7$: $u=0,\sigma=0.5$ &       &       &       &       &        \\
    \multirow{3}[0]{*}{DECNws3} & \multirow{3}[0]{*}{10} & \multirow{3}[0]{*}{2} & \multirow{3}[0]{*}{32} & \multirow{3}[0]{*}{3} & $3\times 3$: $u=0,\sigma=0.5$ & \multirow{3}[0]{*}{0.0005} & \multirow{3}[0]{*}{5000} & \multirow{3}[0]{*}{10} & \multirow{3}[0]{*}{True} & \multirow{3}[0]{*}{10} \\
          &       &       &       &       & $5\times 5$: $u=0,\sigma=0.5$ &       &       &       &       &        \\
          &       &       &       &       & $7\times 7$: $u=0,\sigma=0.5$ &       &       &       &       &        \\
    \multirow{3}[1]{*}{DECNn15} & \multirow{3}[1]{*}{10} & \multirow{3}[1]{*}{30} & \multirow{3}[1]{*}{16} & \multirow{3}[1]{*}{15} & $3\times 3$: $u=0,\sigma=0.5$ & \multirow{3}[1]{*}{0.0005} & \multirow{3}[1]{*}{2000} & \multirow{3}[1]{*}{10} & \multirow{3}[1]{*}{False} & \multirow{3}[1]{*}{10} \\
          &       &       &       &       & $5\times 5$: $u=0,\sigma=0.5$ &       &       &       &       &        \\
          &       &       &       &       & $7\times 7$: $u=0,\sigma=0.5$ &       &       &       &       &        \\
    \bottomrule
    \end{tabular}
\end{center}
\end{table*}

\partitle{Parameters} DECN is compared with standard EA baselines (DE (DE/rand/1/bin) \cite{das2010differential} and CMA-ES). DE are implemented based on Geatpy \cite{geatpy}, LSHADE is implemented by Pyade \footnote{https://github.com/xKuZz/pyade}, and CMA-ES and IPOP-CMA-ES is implemented by cmaes \footnote{https://github.com/CyberAgentAILab/cmaes}. The model parameters of LGA and LES are copyed from evosax \cite{evosax2022github} as suggested by the original paper. For all baselines, the population size is set to 100, and the maximum number of function evaluations is set to 100. We tuned the all parameters in the range of $[0.0, 1.0]$ in steps of 0.05 of all baselines and DECN via a grid sweep. All algorithms are run ten times for each function.

\begin{figure*}[htb]
\centering
\subfloat[F4, $D$=10] {\includegraphics[width=0.33\linewidth]{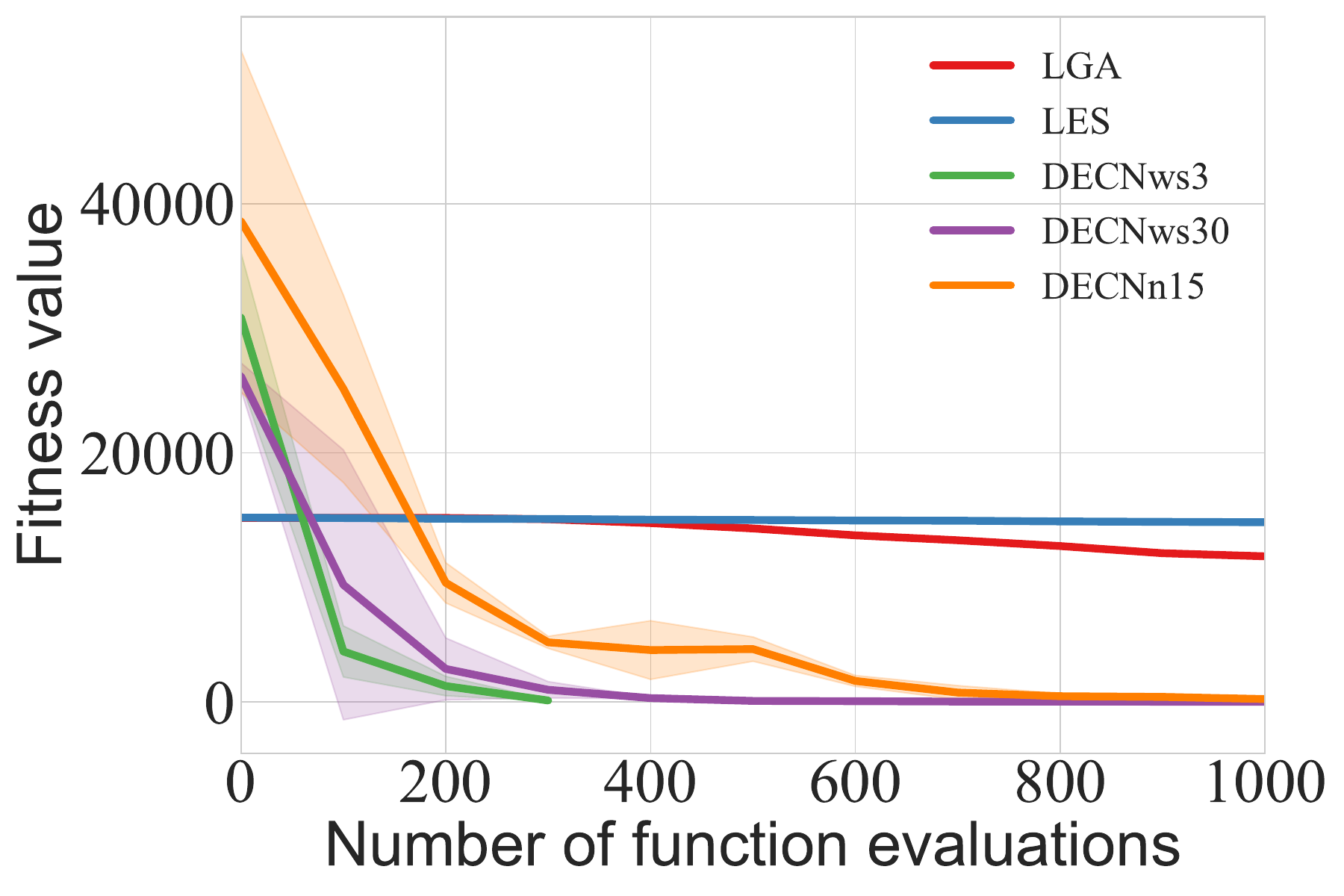}}
%	\hspace{-10pt}
\subfloat[F5, $D$=10] {\includegraphics[width=0.33\linewidth]{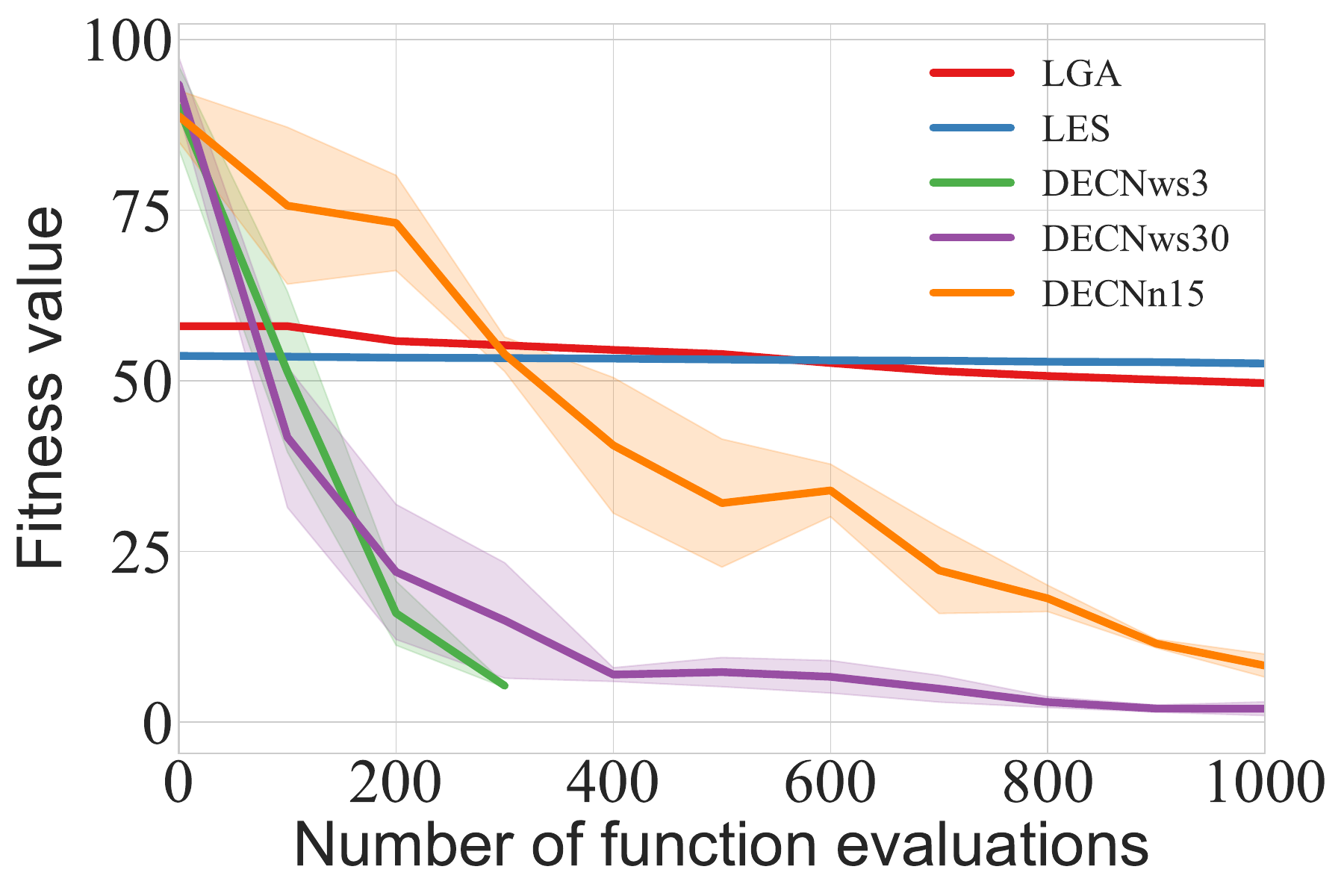}}
\subfloat[F6, $D$=10] {\includegraphics[width=0.33\linewidth]{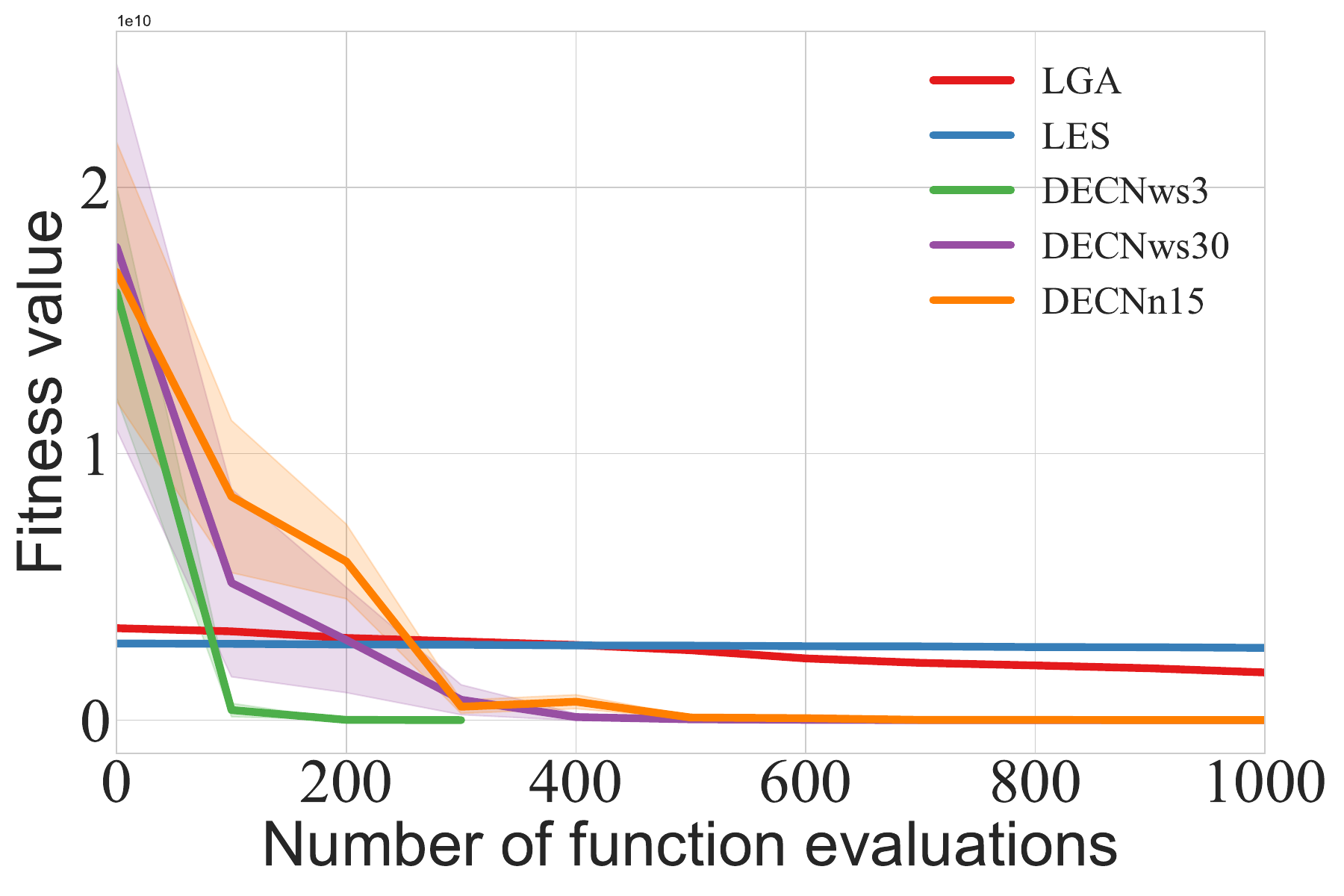}}\\
\subfloat[F7, $D$=10] {\includegraphics[width=0.33\linewidth]{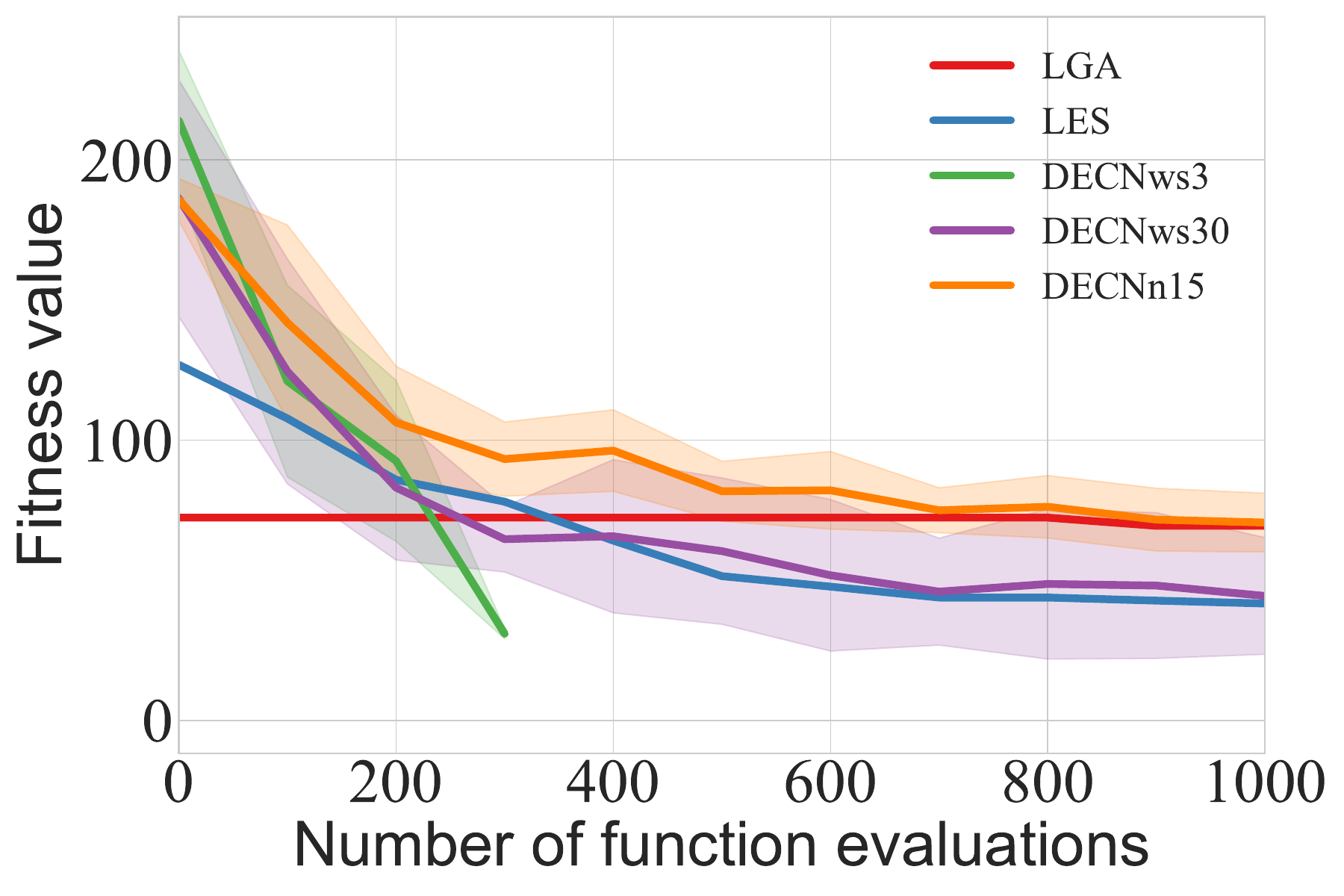}}
%	\hspace{-10pt}
\subfloat[F8, $D$=10] {\includegraphics[width=0.33\linewidth]{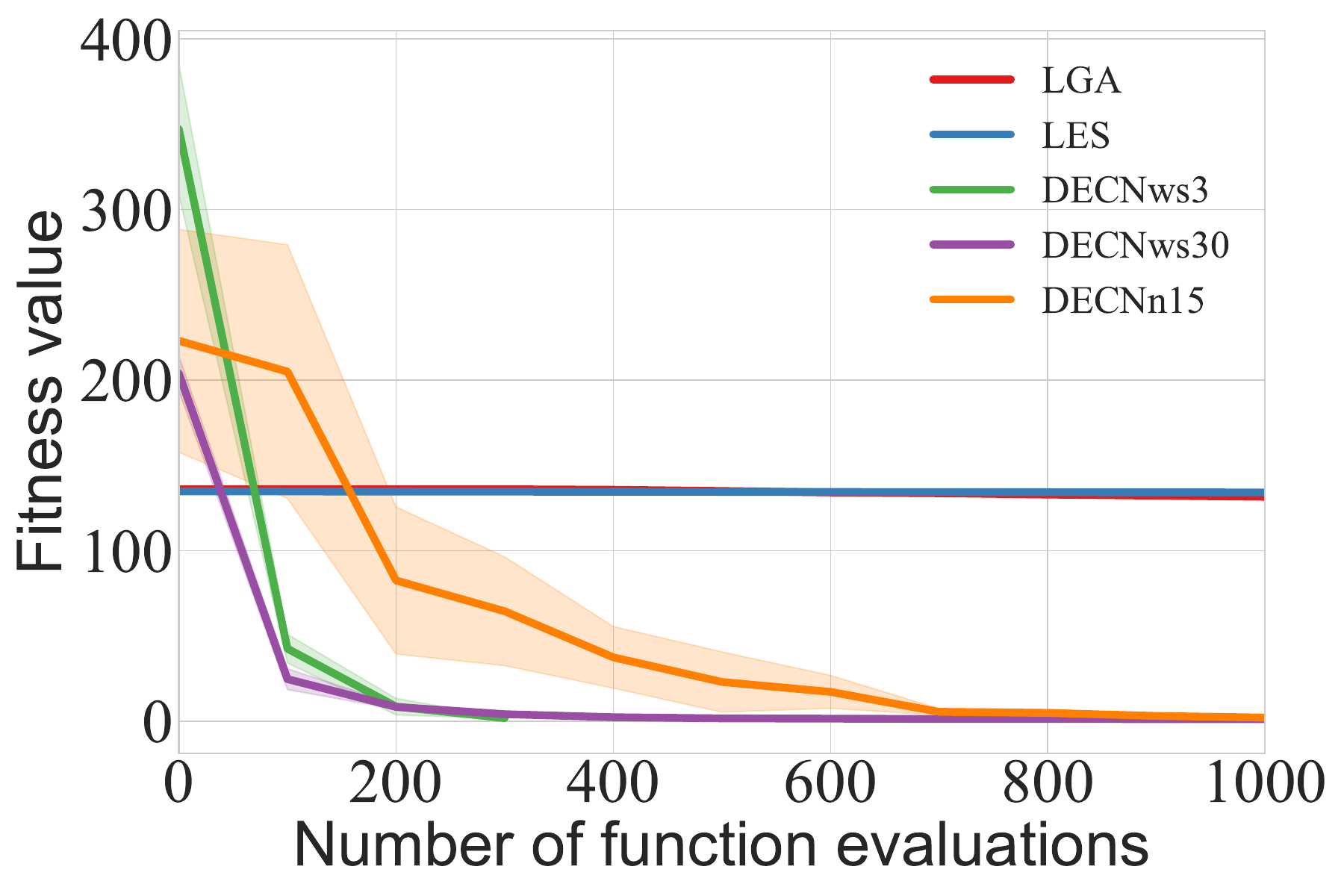}}
\subfloat[F9, $D$=10] {\includegraphics[width=0.33\linewidth]{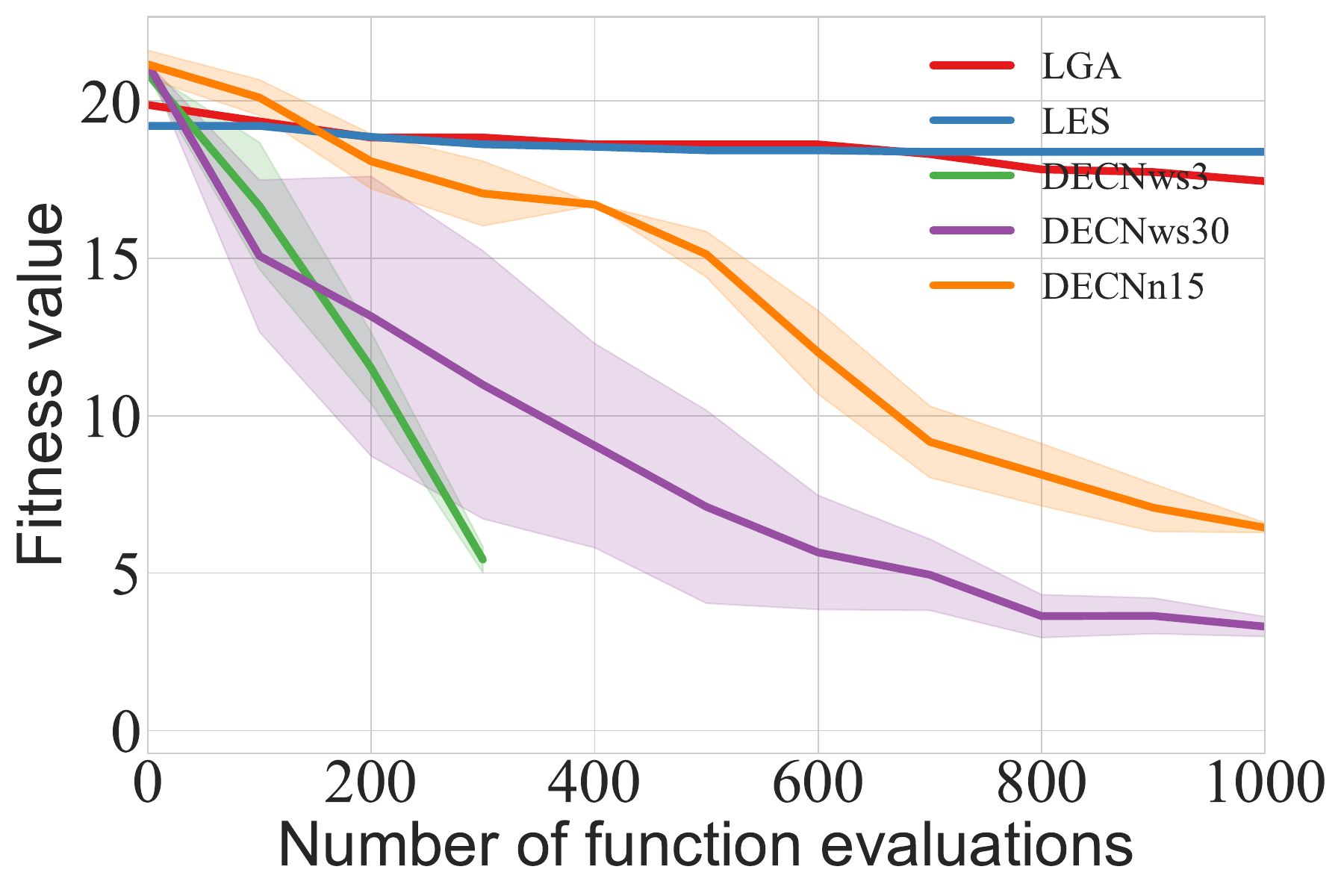}}\\
\caption{The compared results of different DECN structures trained based on low-fidelity surrogate functions. (a)-(f) represent the convergence curves of DECN and the comparison algorithms on F4-F9 when $D$=10.}
\label{fig:r3}
\end{figure*}

\begin{figure*}[htb]
%\vskip -0.2in
\centering
\subfloat[F4, $D$=100] {\includegraphics[width=0.33\linewidth]{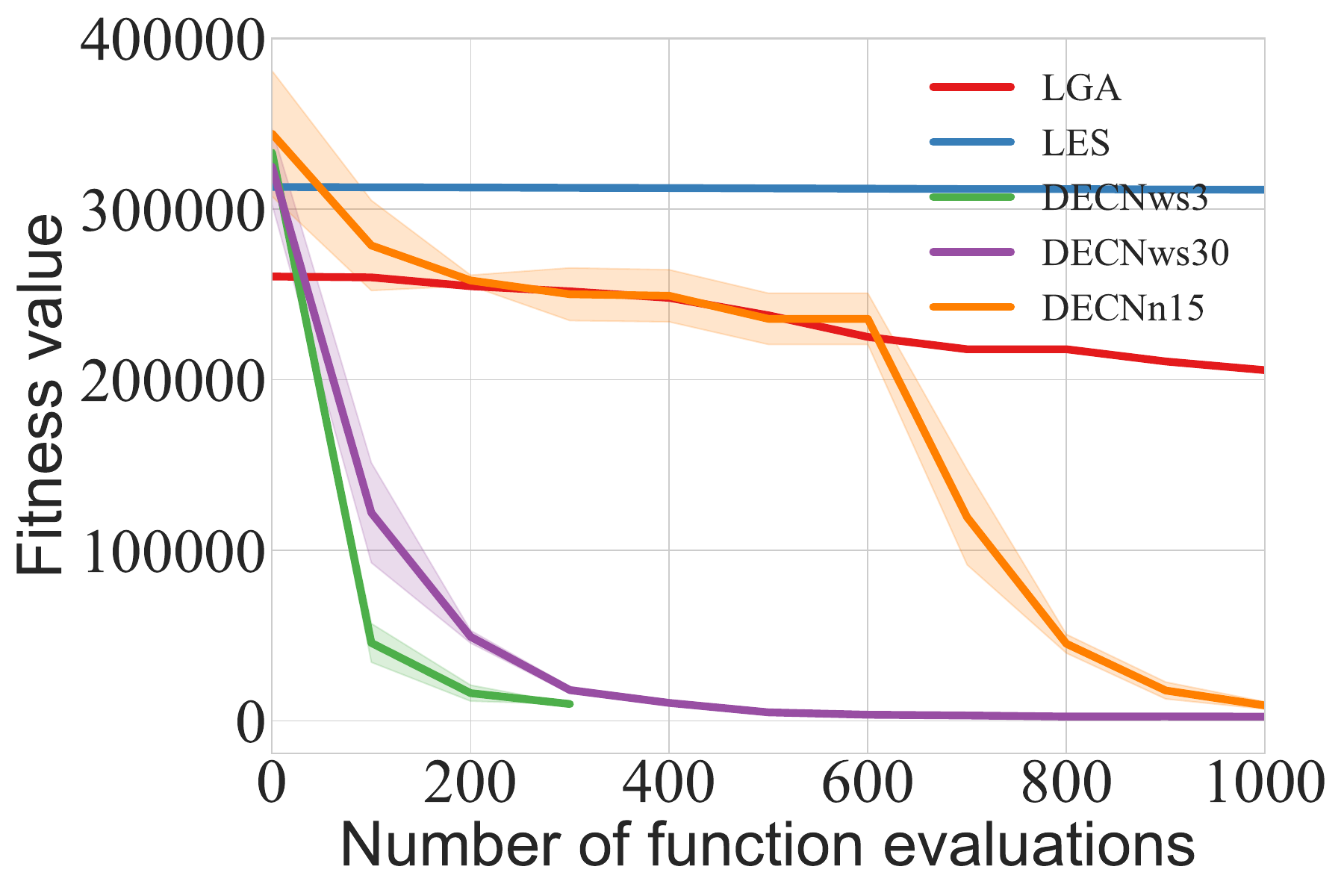}}
%	\hspace{-10pt}
\subfloat[F5, $D$=100] {\includegraphics[width=0.33\linewidth]{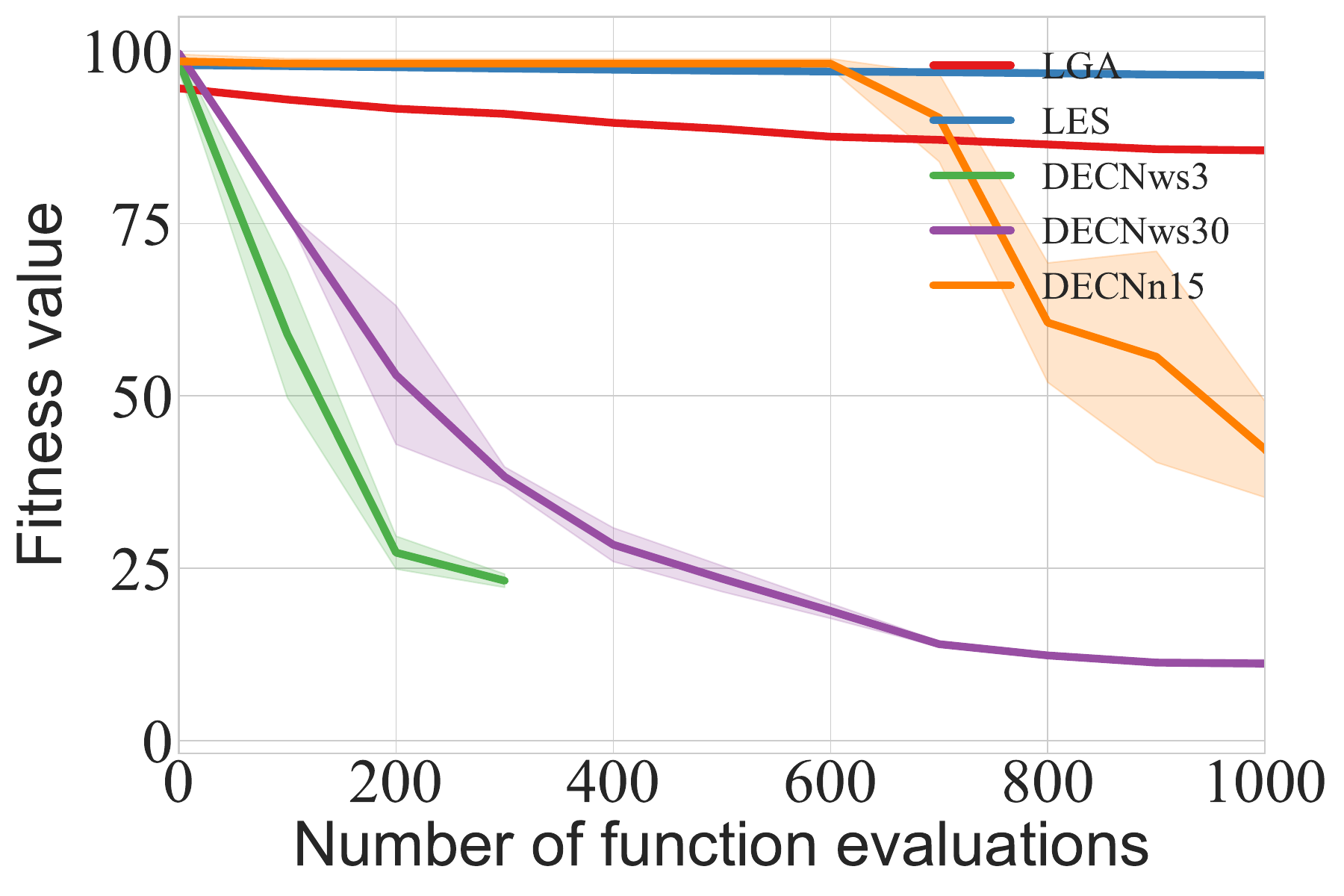}}
\subfloat[F6, $D$=100] {\includegraphics[width=0.33\linewidth]{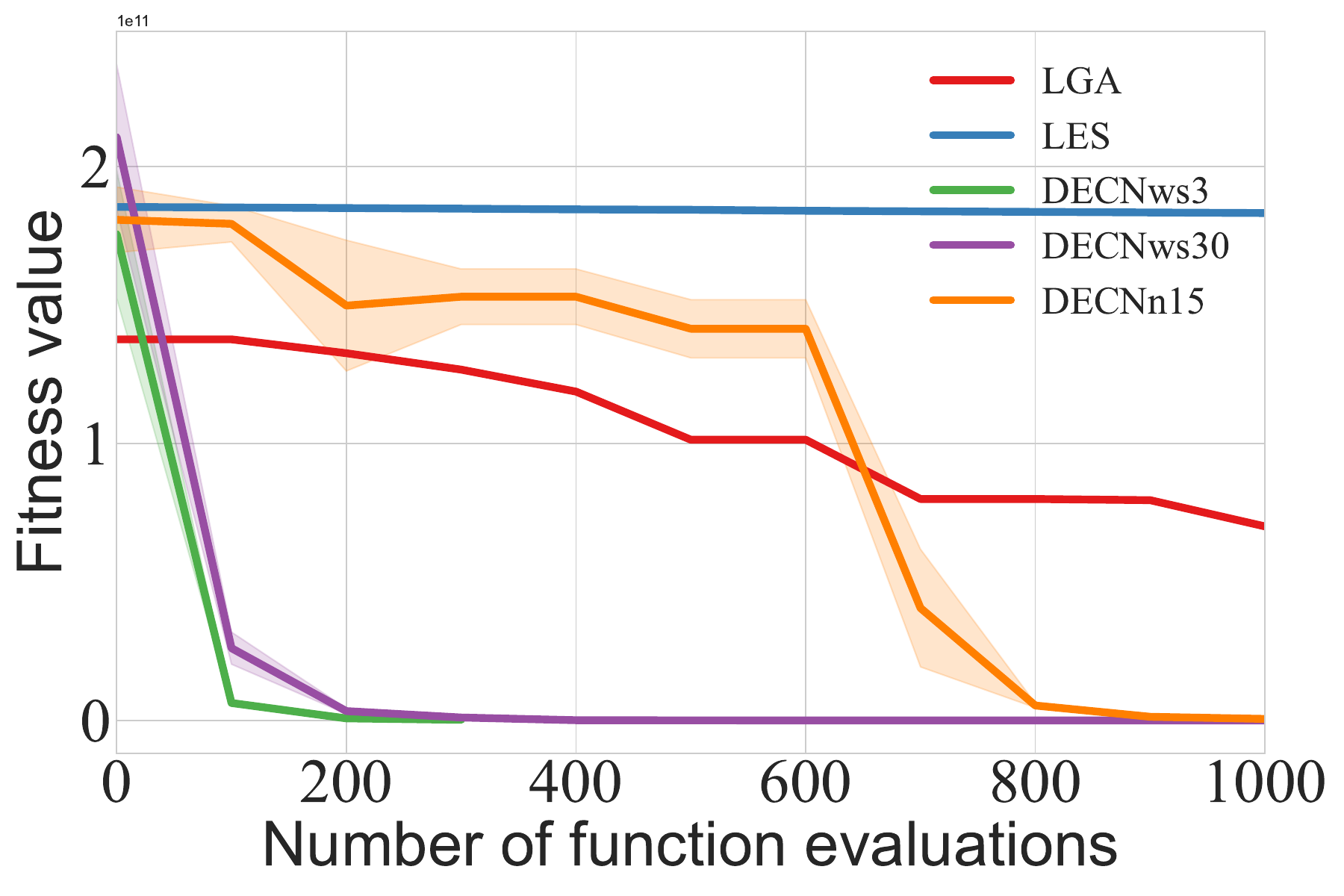}}\\
\subfloat[F7, $D$=100] {\includegraphics[width=0.33\linewidth]{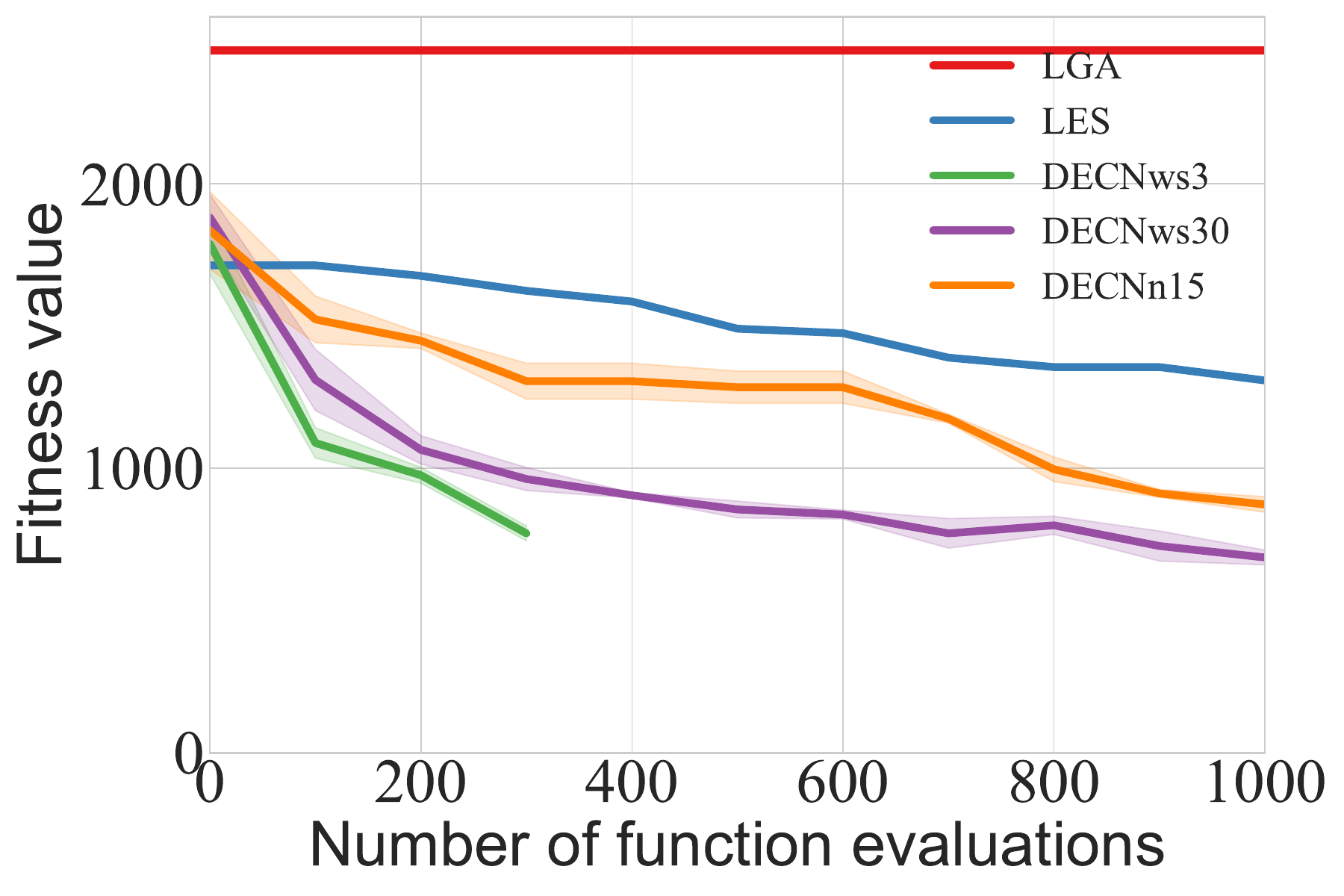}}
%	\hspace{-10pt}
\subfloat[F8, $D$=100] {\includegraphics[width=0.33\linewidth]{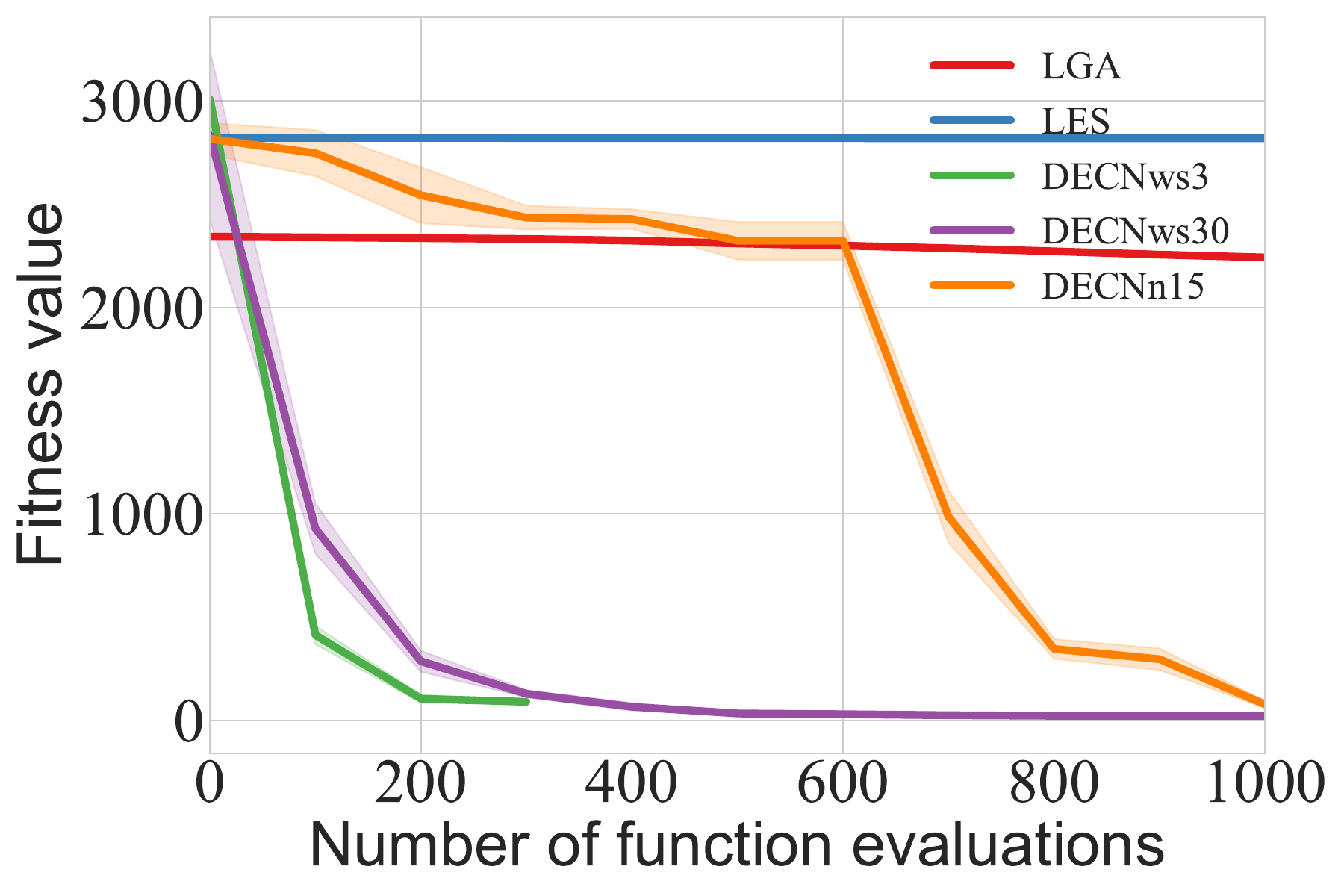}}
\subfloat[F9, $D$=100] {\includegraphics[width=0.33\linewidth]{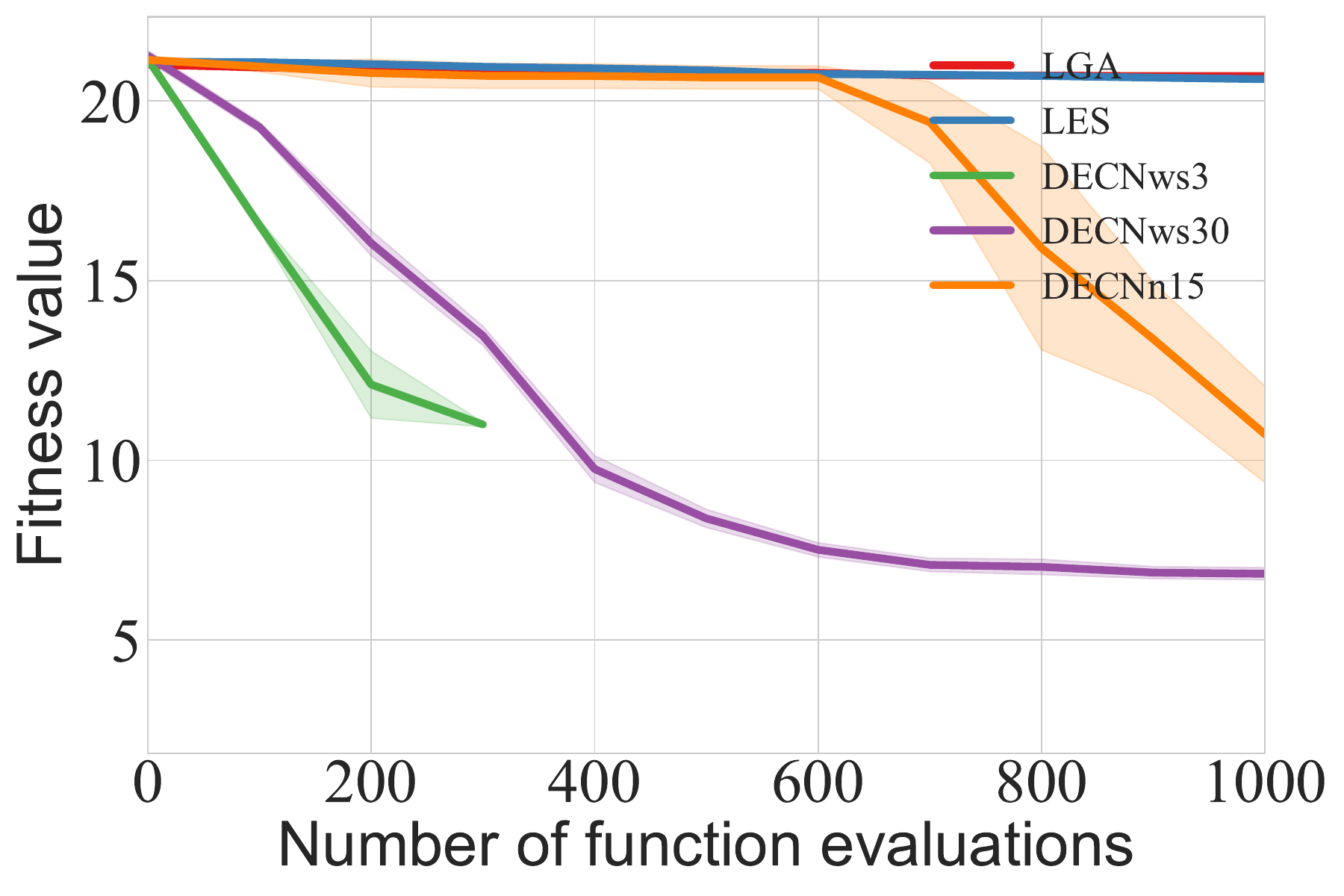}}\\
\caption{The compared results of different DECN structures trained based on low-fidelity surrogate functions. (a)-(f) represent the convergence curves of DECN and the comparison algorithms when $D$=100.}
%\vskip -0.2in
\label{fig:r4}
\end{figure*}

\subsection{Results on Synthetic Functions}
\subsubsection{Results on High-fidelity Training Dataset}

This section validates the strong optimization capabilities imparted by DECN, harnessing the rich high-fidelity information inherent to the target task. Details regarding the generation of high-fidelity surrogate functions can be found as follows, with their assessment conducted on F4-F9 (Table 2).
For each function in Table 2, we produce the training dataset as follows: 1) Randomly initialize the input population $S_0$; 2) Randomly produce a shifted objective function $f_i(s|\xi)$ by adjusting the corresponding location of optima-namely, adjusting the parameter $\xi$; 3) Evaluate $S_0$ by $f_i(s|\xi)$; 4) Repeat Steps 1)-3) to generate the corresponding dataset. For example, we show the designed training and testing datasets for the F4 function as follows:
\begin{equation}\label{eq:hft}
\begin{split}
\mathcal{D} &= \{F4(s|\xi_{1}^{train}),\cdots, F4(s|\xi_{m}^{train})\}, \\
&\mathcal{D}^{test} = \{F4(s|\xi^{test})\}
\end{split}
\end{equation}
$\mathcal{D}$ and $\mathcal{D}^{test}$ are comprised of the same essential function but vary in the location of optima obtained by setting different combinations of $\xi$ (called $b_i$ in Table 1). $\mathcal{D}$ is considered as the high-fidelity surrogate functions of $\mathcal{D}^{test}$. %We train DECN on $F^{train}$, and then we test the performance of DECN upon $F^{test}$, where the values of $\xi^{test}$ not appearing in the training process.

Here, $D=\{10, 100\}$ and $L=10$. 
As depicted in Figs. \ref{fig:r1} and \ref{fig:r2}, a comparative analysis between DECNws3 and state-of-the-art (SOTA) methods is presented. DECN consistently outperforms the compared methods by a substantial margin. This unequivocally underscores DECN's adeptness in effectively leveraging information from the objective function to adapt its optimization strategy. The trained DECN possesses an optimization strategy that is distinctly tailored to the specific task.

\subsubsection{Results on Low-fidelity Training Dataset}

In practice, acquiring high-fidelity surrogate functions for numerous target tasks can prove challenging. This experimental section aims to assess DECN's capacity to navigate such scenarios, specifically focusing on its generalization prowess. To this end, we train DECN using low-fidelity surrogate functions as delineated in Table 1, and subsequently, we subject it to testing on each function enumerated in Table 2. Additionally, we scrutinize the influence of distinct architectures on DECN, encompassing variations in the number of layers and the sharing of weights between layers. The results are visualized in Figs. \ref{fig:r3} and \ref{fig:r4}.

%Here, the whole functions in Table \ref{table:a1} are employed as $F^{train}$ in order to train one DECN, and then the results on each function of Table \ref{table:a2} are shown in Table \ref{table:trans}. 
For example, we show the designed training and testing datasets for the F4 function as follows:
\begin{equation}\label{eq:lft}
\begin{split}
\mathcal{D} &= \{F1(s|\xi_{1,i}), F2(s|\xi_{2,i}), 
 F3(s|\xi_{3,i})\}, \\  
 &\mathcal{D}^{test} = \{F4(s|\xi^{test})\}
\end{split}
\end{equation}
Meanwhile, we also test the impact of different architectures on DECN, including the different number of layers and whether weights are shared between layers. 

Across all instances, DECNws3 exhibits the swiftest convergence rate, followed by DECNws30, while DECNnw15 displays the slowest convergence. These disparities can be attributed to variations in the number of Evolution Modules (EMs) integrated into their respective structures. DECNws3 requires a mere 3 EMs to yield optimal solutions, whereas DECNws30 necessitates 30 EMs for optimal outcomes. For scenarios involving fewer function evaluations, DECNws3 is the recommended choice. Importantly, DECNws30 consistently outperforms DECNws3 in all scenarios, underscoring that deeper architectures possess enhanced representation capabilities, enabling the construction of more precise optimization strategies.

DECNnw15 surpasses DECNws3 and DECNws30 in most cases when the number of function evaluations increases (refer to \textit{SI appendix, Table S4}). Despite DECNn15 having fewer layers than DECNws30, its representation capacity proves superior due to its non-weight-sharing nature. However, it's worth noting that as the number of layers increases, this architecture becomes more challenging to train effectively.
Moreover, DECN outperforms LES and LGA, serving as evidence of DECN's superior generalization and optimal policy characterization abilities.

\subsubsection{Discussion}
\partitle{Factors Underlying Strong Performance}
We trained DECN using F1-F3 (Table 1) and subsequently applied it successfully to F4-F9. DECN's superior performance over LES and LGA is indicative of its acquisition of more generalized optimization expertise. This is attributed to the following factors:

\textit{1) Insights from Function Expressions}: F1-F3 encompass crucial information related to the objective functions found in F4-F9. For instance, F7 can be decomposed into $\sum_{i=1}^{D} z_i^2-\sum_{i=1}^{D}10\cos(2\pi z_i)+ \sum_{i=1}^{D}10$. F2 serves as the low-fidelity surrogate function of $\sum_{i=1}^{D} z_i^2$, while F1 represents the low-fidelity surrogate function of $\sum_{i=1}^{D}10\cos(2\pi z_i)$. Similar surrogate functions from F1-F3 can be identified for other functions in F4-F9. DECN capitalizes on this information to maximize the alignment between its learned optimization strategy and the objective function. However, it's important to note that F6 exhibits less resemblance to F1-F3 than F4, F5, and F7-F9. Consequently, while DECN performs better on F6 than the baseline methods, there is still room for improvement.

\textit{2) Landscape Characteristics}: F1-F3 incorporate a diverse set of landscape features, including unimodal, multimodal, separable, and non-separable characteristics. The landscape features present in F4-F9 are as follows: 1) F4, F5: Unimodal, Separable;
2) F6: Multimodal, Non-separable, Having a very narrow valley from local optimum to global optimum, ill-conditioned;
3)F7: Multimodal, Separable, Asymmetrical, Local optima’s number is huge;
4) F8: Multi-modal, Non-separable, Rotated;
5) F9: Multi-modal, Non-separable, Asymmetrical.
These differences in landscape features highlight DECN's adaptability and capacity to tackle a wide range of optimization scenarios.

The landscape features present in F4 and F5 share similarities with those found in F1-F3, indicating some degree of overlap. In contrast, F6-F9 introduce novel features. The varying impact of these distinct characteristics on the landscape can be ranked as follows: Having a very narrow valley from local optimum to global optimum, ill-conditioned$>$Asymmetrical, Local optima’s number is huge$>$Asymmetrical$>$Rotated. Consequently, DECN exhibits the highest level of generalization performance on F4, F5, and F8, a moderately strong performance on F7 and F9, and its performance is least optimal on F6.

\begin{figure*}[htbp]
\begin{center}
\centerline{\includegraphics[width=1\linewidth]{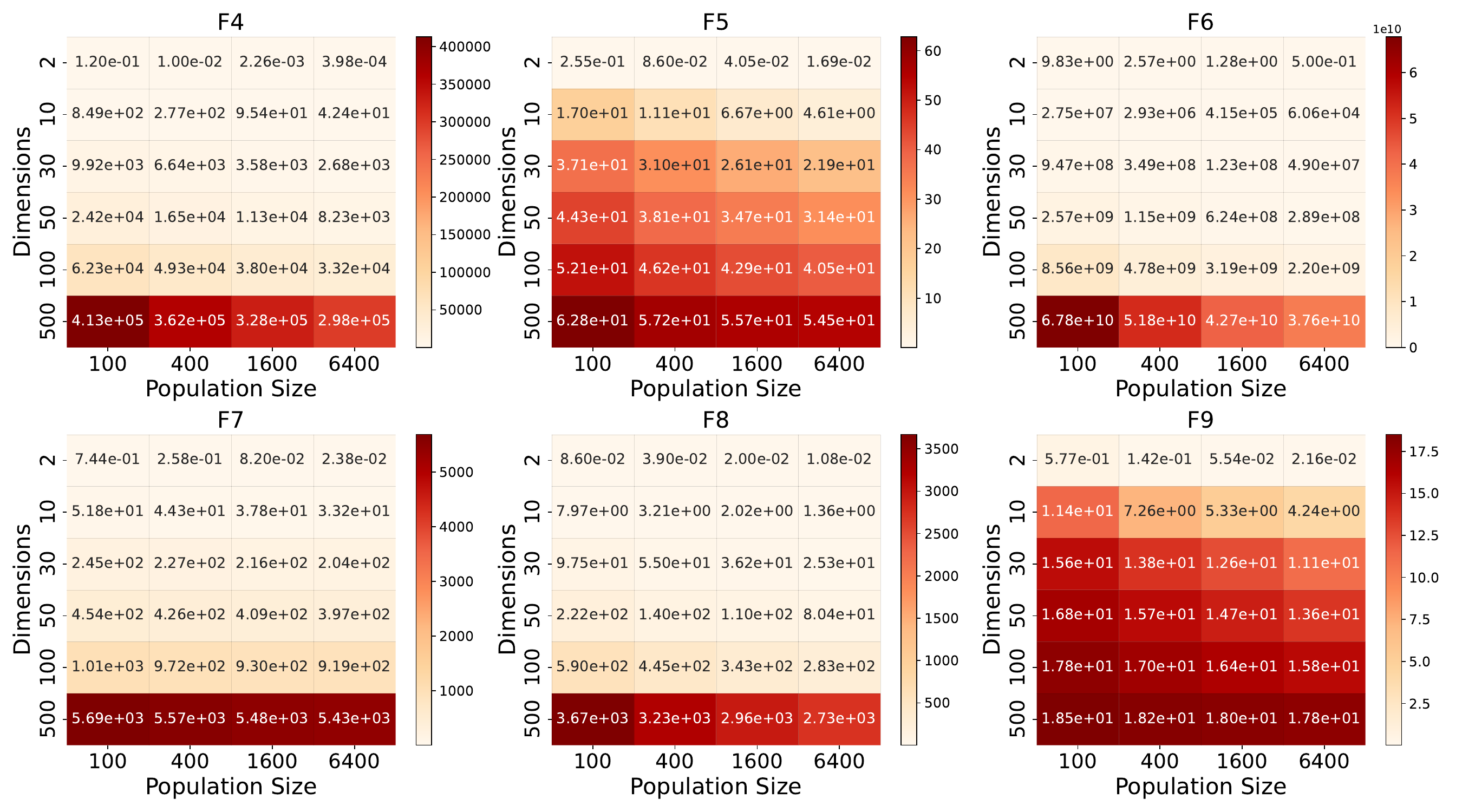}}
\caption{The performance of DECNws3 with different $L$ and $D$. Our training dataset $\mathcal{D}$, crafted with $D=2$ and $L=10$, is founded on functions detailed in Table 1. Subsequently, we subject DECN to evaluation using the functions outlined in Table 2.}
\label{fig:DL}
\end{center}
\end{figure*}

The performance of DECN demonstrates a correlation with the similarity of fitness landscapes between the training set and the specific problem at hand. Even when novel problem attributes not present in the training set are encountered, DECN maintains a strong performance. However, when the training dataset lacks extreme attributes found in certain problems, DECN's performance may be less optimal for functions characterized by such attributes. These results underscore the generality and transferability of the optimization strategy acquired by DECN to a wide range of previously unseen objective functions.

\partitle{Effect of Training Dataset} 
DECN's performance is inherently linked to the quality of its training dataset, a common concern in deep learning frameworks. Notably, even when trained on a low-fidelity dataset, DECN consistently outperforms LES and LGA, two state-of-the-art meta-learning EAs, despite the training dataset for LES/LGA being of significantly higher quality.

\partitle{Generalization capability of DECN} 
%We have validated DECN's generalization capability (refer to Section 4.2). DECN was trained on both high-fidelity and low-fidelity datasets for the target task, revealing the following insights: 1) DECN, trained on the high-fidelity dataset, outperforms state-of-the-art human-designed evolutionary algorithms and meta-learning EAs. This observation signifies DECN's adeptness at generalizing to the target task, although its ability to generalize to non-target tasks is comparatively limited. 2) DECN, trained on the low-fidelity dataset, surpasses state-of-the-art meta-learning EAs, demonstrating robust generalization capabilities.
We extensively validated DECN's generalization capacity, as detailed in Section 4.2. Our findings reveal the following key insights: 1) DECN, trained on high-fidelity data, surpasses both state-of-the-art human-designed evolutionary algorithms and meta-learning EAs. This underscores DECN's strong generalization within the target task domain, although its ability to generalize to non-target tasks is comparatively limited. 2) DECN, when trained on low-fidelity data, consistently outperforms state-of-the-art meta-learning EAs, showcasing its robust generalization capabilities.

\partitle{Efficiency of DECN Scales with Complexity and Size} 

We assessed DECN's scalability concerning problem complexity and size, as elaborated in the "Results on Synthetic Functions" section. Our main findings are summarized below:

DECN's performance was examined using the high-fidelity training dataset, yielding the following observations. Problem difficulty ranks as follows: F7$=$F8$<$F4$<$F5$=$F9$\ll$F6. It becomes evident that DECN's performance tends to decrease with increasing problem complexity. The comparison between dimensions $D=10$ and $D=100$ further highlights DECN's sensitivity to problem dimensionality.

We also evaluated DECN's performance with the low-fidelity training dataset. In this context, DECN's performance shows a consistent decline as problem complexity increases. Comparing results across dimensions $D=10$ and $D=100$ underscores DECN's performance decrease with higher problem dimensionality. To enhance DECN's performance, strategies such as increasing its depth (e.g., DECNws30) or avoiding weight-sharing approaches can be employed.

\subsubsection{DECN's Generalization Across Diverse Optimization Scenarios}

We extensively validate DECN's generalization capacity across various optimization scenarios, encompassing distinct tasks, populations with varying dimensions ($D$), and different scales ($L$). The results for F4-F9 are illustrated in Fig. \ref{fig:DL}. Notably, DECN exhibits commendable performance, underscoring its capacity to accumulate broad optimization expertise.

This tested DECNws3 model, initially trained within a 2-dimensional optimization environment, maintains good performance even as it confronts high-dimensional problems. An intriguing observation is that its performance is further enhanced with larger values of $L$, as a greater number of individuals bolster DECN's search capabilities. Conversely, as $D$ increases, DECN's performance dips, primarily due to the diminishing similarity between the training set and the target task.

\subsubsection{GPU Acceleration and Performance Comparison}
Leveraging GPU acceleration is often a challenging endeavor for EAs. The logical operations inherent in many genetic operators are not inherently conducive to GPU processing. However, in the case of DECN, both the CRM and SM predominantly involve tensor operations, lending themselves to efficient GPU acceleration.

In Table \ref{table:a4}, we present a comprehensive runtime comparison between DECNws3 and traditional EA. The results manifestly demonstrate the advantages of GPU-based acceleration, particularly as the population size ($L$) increases. For $D$ values in the range of $2$ to $10$, DECN outperforms EA by a substantial factor, falling within the range of 103 to 104 times faster. Even with more demanding scenarios where $D$ spans from $30$ to $500$, DECN maintains an approximate 102-fold speed advantage over EA. This vividly underscores DECN's compatibility with GPU acceleration and the remarkable performance enhancements it can provide. Notably, as $D$ increases, DECN's runtime allocation to evaluations grows, tempering the acceleration advantage. This observation highlights the proficiency of DECN in GPU acceleration, particularly when optimizing larger populations.

\begin{table}[htbp]
\caption{We explore the computational efficiency of DECN when accelerated using a single 1080Ti GPU. The values presented in this table represent the average time in seconds that each algorithm requires to perform three generations of evolution for 32 input populations. Notably, the EA employs a combination of SBX crossover, Breeder mutation, and binary tournament mating selection operators.}
\label{table:a4}
\begin{center}
    \begin{tabular}{ccrrrr}
    \toprule
    \multirow{2}[3]{*}{D} & \multirow{2}[3]{*}{Algorithm} & \multicolumn{4}{c}{\textit{L}} \\
\cmidrule{3-6}          &       & \multicolumn{1}{c}{10} & \multicolumn{1}{c}{20} & \multicolumn{1}{c}{40} & \multicolumn{1}{c}{80} \\
    \multirow{3}[1]{*}{2} & DECN(s) & 0.005 & 0.006 & 0.007 & 0.015 \\
          & EA(s) & 0.700 & 2.863 & 12.676 & 71.740 \\
          & Rate(DECN/EA) & 0.0066 & 0.002 & 0.0006 & 0.0002 \\
    \midrule
    \multirow{3}[2]{*}{10} & DECN  & 0.0055 & 0.0078 & 0.017 & 0.050 \\
          & EA    & 0.694 & 2.880 & 12.876 & 72.366 \\
          & Rate(EM/EA) & 0.0079 & 0.0027 & 0.0013 & 0.0007 \\
    \midrule
    \multirow{3}[2]{*}{30} & DECN  & 0.007 & 0.013 & 0.040 & 0.139 \\
          & EA    & 0.693 & 2.879 & 13.004 & 71.675 \\
          & Rate(EM/ EA) & 0.010 & 0.0046 & 0.0031 & 0.0019 \\
    \midrule
    \multirow{3}[2]{*}{50} & DECN  & 0.008 & 0.0188 & 0.062 & 0.237182 \\
          & EA    & 0.682 & 2.868 & 12.807 & 71.442 \\
          & Rate(EM/ EA) & 0.012 & 0.0065 & 0.0048 & 0.003 \\
    \midrule
    \multirow{3}[2]{*}{100} & DECN  & 0.0117 & 0.033 & 0.117 & 0.479 \\
          & EA    & 0.699 & 2.866 & 13.122 & 71.830 \\
          & Rate(EM/ EA) & 0.0168 & 0.0116 & 0.009 & 0.0067 \\
    \midrule
    \multirow{3}[2]{*}{500} & DECN  & 0.041 & 0.148 & 0.610 & 2.727 \\
          & EA    & 0.721 & 2.967 & 13.600 & 74.942 \\
          & Rate(EM/ EA) & 0.057 & 0.050 & 0.045 & 0.036 \\
    \bottomrule
    \end{tabular}%
\end{center}
\end{table}

\subsection{Results on Robotic Control Task}
%We discuss how to construct training datasets on real-world tasks. Also, the effectiveness of DECN is verified in two important scenarios.

The planner mechanic arm problem \cite{Cully2015RobotsTC,Mouret2020QualityDF} is important in robotics. This problem is to find the suitable sets of angles ($\alpha \in R^D$) and lengths ($\beta \in R^D$) such that the distance $f(L, \alpha, p)$ from the top of the mechanic arm to the target position $p$ is the smallest. The planner mechanic arm has been frequently employed as an optimization problem to assess how well the black-box optimization algorithms perform \cite{Cully2015RobotsTC,Vassiliades2018UsingCV,Vassiliades2018DiscoveringTE,Mouret2020QualityDF}. The planner mechanic arm problem has two key parameters: the set of $L=(L_1, L_2, \cdots, L_n)$ and the set of angles $\alpha=(\alpha_1, \alpha_2, \cdots, \alpha_n)$, where $n$ represents the number of segments of the mechanic arm, and $L_i \in (0, 10)$ and $\alpha_i \in (-\Pi, \Pi)$ represent the length and angle of the $i$th mechanic arm, respectively. This problem is to find the suitable sets of $L$ and $\alpha$ such that the distance $f(L, \alpha, p)$ from the top of the mechanic arm to the target position $p$ is the smallest, where $f(L, \alpha, p) = \sqrt{\left(\sum_{i=1}^n \cos(\alpha_i)L_i-p_x\right)^2 + \left(\sum_{i=1}^n \sin(\alpha_i)L_i-p_y\right)^2}$, and $(p_x, p_y)$ represents the target point's x- and y-coordinates. Here, $n=100$. We design two groups of experiments.

\begin{figure}[htbp]
\begin{center}
\centerline{\includegraphics[width=0.85\columnwidth]{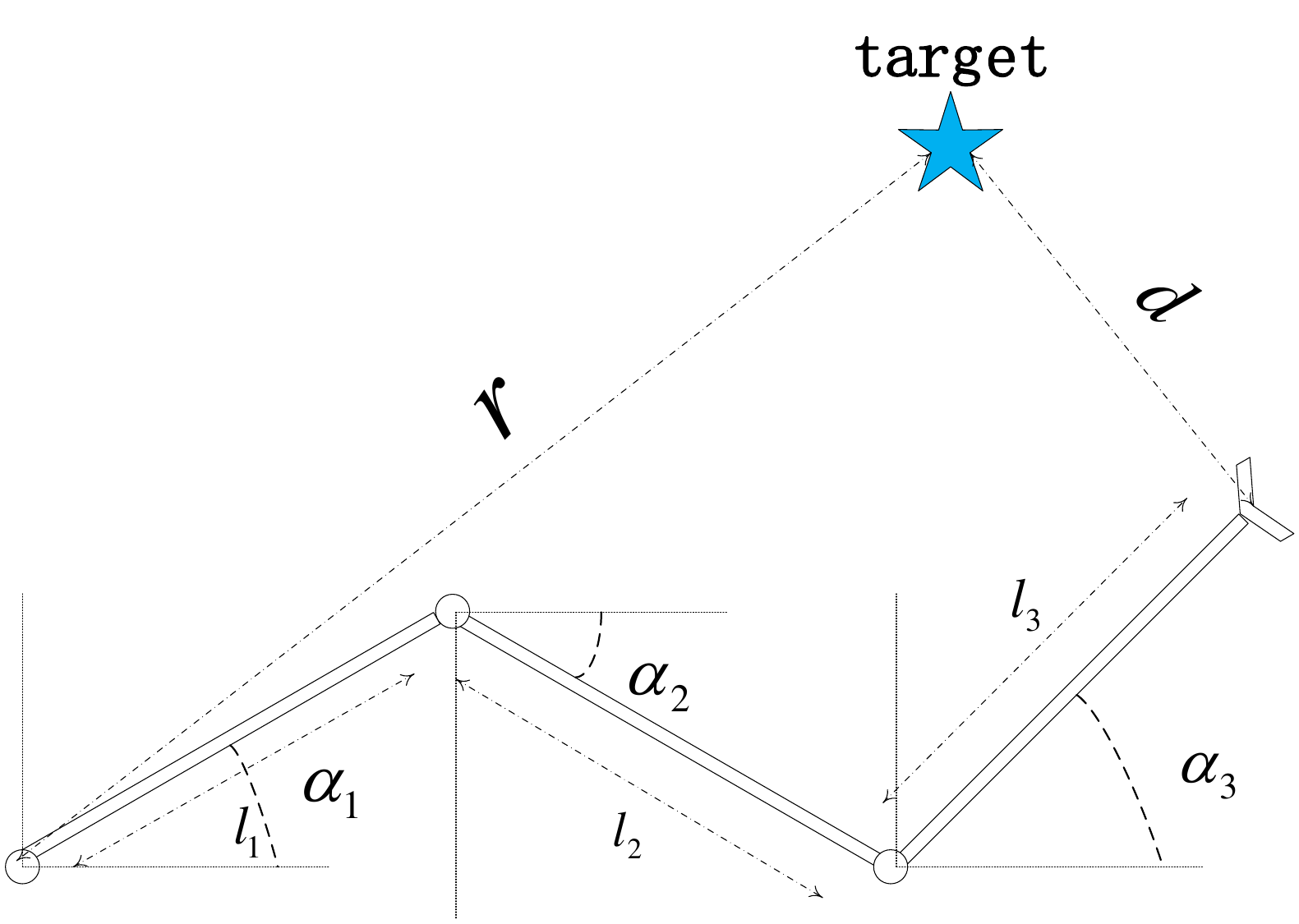}}
\caption{Planar Mechanical Arm.}
\label{fig:a7}
\end{center}
\end{figure}

\begin{table*}[!t]
  \centering
  \caption{The results of planar mechanical arm. $FE$ is the number of function evaluations. Here, *(*) represents the mean (standard deviation).}
  \vskip -0.15in
       \begin{tabular}{ccccccccc|c}
\toprule
\multicolumn{1}{c}{Case} & $FE$    & $r$     & DE    & CMA-ES & L-SHADE & I-POP-CMA-ES & LES   & LGA   & DECN \\
\midrule
\multirow{6}[12]{*}{SC} & \multirow{2}[4]{*}{1000} & 100   & \multicolumn{1}{c}{2.96(1.63)} & \multicolumn{1}{c}{7.19(2.09)} & 2.65(0.13) & 2.11(0.77) & 0.92(0.47) & 3.76(1.93) & \textbf{0.42(0.22)} \\
&       & 300   & \multicolumn{1}{c}{11.3(14.7)} & \multicolumn{1}{c}{10.9(6.92)} & 3.41(0.15) & 2.42(0.89) & 2.40(5.05) & 25(29.3) & \textbf{1.04(1.25)} \\
\cmidrule{2-10}          & \multirow{2}[4]{*}{2000} & 100   & 1.28(0.60) & 6.87(2.12) & 1.12(0.71) & 1.19(0.46) & 0.35(0.17) & 1.58(0.85) & \textbf{0.28(0.14)} \\
&       & 300   & 1.54(0.89) & 7.55(2.36) & 1.30(0.06) & 1.28(0.45) & 0.38(0.19) & 13.8(20) & \textbf{0.30(0.11)} \\
\cmidrule{2-10}          & \multirow{2}[4]{*}{3000} & 100   & 1.20(0.64) & 6.12(2.00) & 0.57(0.05) & 0.69(0.29) & 0.22(0.12) & 1.09(0.62) & \textbf{0.14(0.12)} \\       &       & 300   & 1.38(0.71) & 6.58(2.36) & 0.65(0.04) & 0.77(0.29) & 0.20(0.10) & 7.70(12.1) & \textbf{0.10(0.09)} \\
\cmidrule{1-10}    \multirow{2}[4]{*}{CC} & \multirow{2}[4]{*}{3000} & 100   & 0.81(0.47) & 4.25(1.20) & 0.29(0.11) & 0.56(0.22) & 0.33(0.17) & 2.55(1.37) & \textbf{0.24(0.15)} \\
&       & 300   & 6.15(12.2) & 5.10(1.71) & 0.51(0.89) & 0.64(0.24) & 0.33(0.18) & 3.35(1.72) & \textbf{0.25(0.13)} \\
    \bottomrule
    \end{tabular}
  \label{tab:pma1}
\end{table*}

\textit{1) Simple Case (SC)}. We fixed the length of each mechanic arm as ten and only searched for the optimal $\alpha$. We randomly selected 600 target points within the range of $r \leq 1000$, where $r$ represents the distance from the target point to the origin of the mechanic arm. In the testing process, we extracted 128 target points in the range of $r \leq 100$ and $r \leq 300$, respectively, for testing. We show the designed training and testing datasets as follows:
\begin{equation}\label{eq:pmaf}
\begin{split}
\mathcal{D} = \{f(\alpha|p_{1}), \cdots, f(\alpha|p_{600})\}, \\
\mathcal{D}^{test} = \{f(\alpha|p_{1}^{test}), \cdots, f(\alpha|p_{128}^{test})\}
\end{split}
\end{equation}
%where $\xi = (p_x, p_y)$.}

\textit{2) Complex Case (CC).} We search for $\beta$ and $\alpha$ at the same time. We show the training and testing datasets as follows:
\begin{equation}\label{eq:pmaf2}
\begin{split}
\mathcal{D} = \{f((\beta,\alpha)|p_{1}),\cdots,f((\beta,\alpha)|p_{600})\}, \\
\mathcal{D}^{test} = \{f((\beta,\alpha)|p_{1}^{test}),\cdots,f((\beta,\alpha)|p_{128}^{test})\}
\end{split}
\end{equation}
The experimental results are presented in Table \ref{tab:pma1}. To ensure fairness, we constrain the number of function estimates used by the DECN architecture to be no higher than that of the comparison algorithms. Specifically, DECNws3 is utilized when $FE=1000$, DECNws15 when $FE=2000$, and DECNws30 when $FE=3000$. Across all scenarios, DECN consistently outperform all baseline methods. Notably, even in scenarios with a broader search space, the performance of DECN exhibits only a slight reduction, yet it still outperforms the comparison algorithms.

%\begin{figure}[htbp]
%\begin{center}
%\centerline{\includegraphics[width=0.9\linewidth]{minist.pdf}}
%\caption{Misclassification accuracy of DECNws30 and LES on MNIST dataset.}
%\label{fig:vem}
%\end{center}
%\end{figure}

\subsection{Results on Training CNN on MNIST Dataset}
DECNws30 is trained based on Table \ref{table:a1} with three functions.
During its training phase, the objective of DECNws30 was to minimize the cross-entropy loss of a convolutional neural network (CNN), whose structure is shown in Table \ref{tab:mnist net}, serving as a surrogate for metric accuracy. In contrast, during the testing phase, DECNws30 and other baseline algorithms aimed to maximize the accuracy of the training set. To create surrogate problems with varying fidelity levels, we utilized 25\% and 50\% of the training data. The results, presented in Table \ref{tab:nn}, consistently demonstrate DECN's superior performance across all fidelity levels.
\begin{table}[htbp]
  \centering
  \small
  \caption{The small-scale classification neural network for MNIST.}
  \resizebox{0.5\textwidth}{!}{
    \begin{tabular}{ccccc}
    \toprule
    Layer ID & Layer Type & Padding & Stride & Kernel Size \\
    \midrule
    1     & depth-wise convolution & \XSolidBrush & 1 & 1$\times$5$\times$5 \\
    2     & point-wise convolution & \XSolidBrush &1 & 3$\times$1$\times$1$\times$1 \\
    3     & ReLu  & \XSolidBrush  & \XSolidBrush  & \XSolidBrush \\
    4     & max pooling & \XSolidBrush & 2 & 2$\times$2 \\
    5     & depth-wise convolution & \XSolidBrush & 1 & 3$\times$5$\times$5 \\
    6     & point-wise convolution & \XSolidBrush & 1 & 16$\times$3$\times$1$\times$1 \\
    7     & ReLu  & \XSolidBrush  & \XSolidBrush  & \XSolidBrush \\
    8     & max pooling & \XSolidBrush & 2 & 2$\times$2 \\
    9     & depth-wise convolution & \XSolidBrush & 1 & 16$\times$4$\times$4 \\
    10    & point-wise convolution & \XSolidBrush & 1 & 10$\times$16$\times$1$\times$1 \\
    11    & softmax & \XSolidBrush  & \XSolidBrush  & \XSolidBrush \\
    \bottomrule
    \end{tabular}}
  \label{tab:mnist net}
\end{table}

\begin{table*}[htbp]
  \centering
  \caption{The classification accuracy of all methods on the MNIST dataset. Datasize represents the proportion of data sets involved in training.}
       \begin{tabular}{ccccccccc|c}
\toprule
Datasize & DECN & CMA-ES & I-POP-CMA-ES& L-SHADE&LGA &LES &\\
\midrule
0.25 &\textbf{0.61(0.01)} & 0.57(0.04)&0.33(0.02)&0.31(0.03)&0.53(0.02)&0.14(0.02)\\
0.5&\textbf{0.70(0.02)} &0.61(0.03)&0.37(0.04)&0.29(0.01)&0.51(0.01)&0.25(0.03)\\
    \bottomrule
    \end{tabular}
  \label{tab:nn}
\end{table*}

%We show that DECNws30 can also successfully be applied to MNIST image classification task with a small CNN with 567 evolvable weights. The detailed structure of CNN can be found in Appendix. 
%DECNws30 to 100 represents the stacking of EMs in DECNws30 repeatedly into 100 layers. The learned EM has some exploration ability.

\subsection{Visualization of DECN}
\begin{figure*}[t]
%\vskip 0.2in
\begin{center}
\centerline{\includegraphics[width=0.75\linewidth]{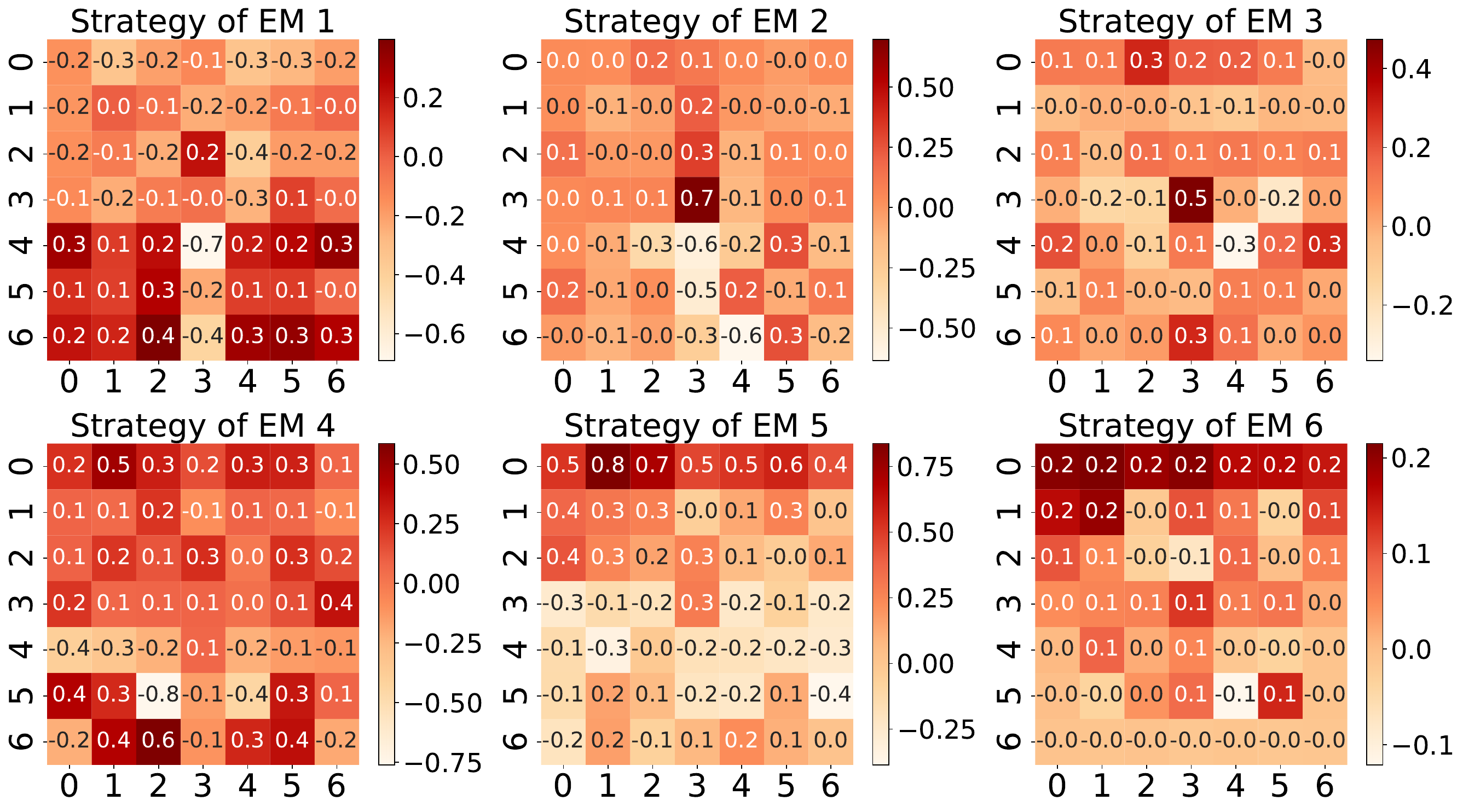}}
\caption{Visualization of $7 \times 7$ convolution kernel. DECNnws6 was trained using the data from Table \ref{table:a2}}
\label{fig:vis}
\end{center}
%\vskip -0.2in
\end{figure*}

We visualize and analyze the optimization strategies learned by DECNnws6. DECNnws6 is trained on Table 1. The learnable parameters of DECN are the convolution kernel of the CRM module. Therefore, we choose a $7 \times 7$ convolution kernel for display, which is shown in Fig. \ref{fig:vis}. The convolution kernel represents the solution generation strategy learned by DECN. From EM1 to EM6, the focus area of the optimization strategy is constantly changing. EM1 focuses on the poor and promising areas in terms of function fitness, moves to the middle area (EM2), and then gradually moves to the area with the better function values. Therefore, DECN first focuses on "exploration" and then gradually shifts to "exploitation". The transformation of the focus of the optimization strategy reflects the dynamic balance between "exploration" and "exploitation" of DECN and proves the intelligence of DECN, which is an advantage that current EAs does not have.

Concurrently, the variability among target tasks engenders corresponding variations within the DECN algorithm. In such instances, DECN's optimization strategies may transcend conventional norms, while remaining grounded in reason and tailored to the objectives of the task at hand, yielding favorable outcomes at the very least. Furthermore, the assumption positing that "the individuals near the good-performing individuals may be better" is both biased and imprecise. Should the good-performing individuals cluster around local optima, it has the potential to mislead subsequent optimization efforts. These biased assumptions collectively pose significant barriers to the further progression of evolutionary computation.

\begin{figure}[ht]
%\vspace{-20pt}
\begin{center}
%\centerline{\includegraphics[width=0.8\linewidth]{visualization.pdf}}
\centerline{\includegraphics[width=0.95\linewidth]{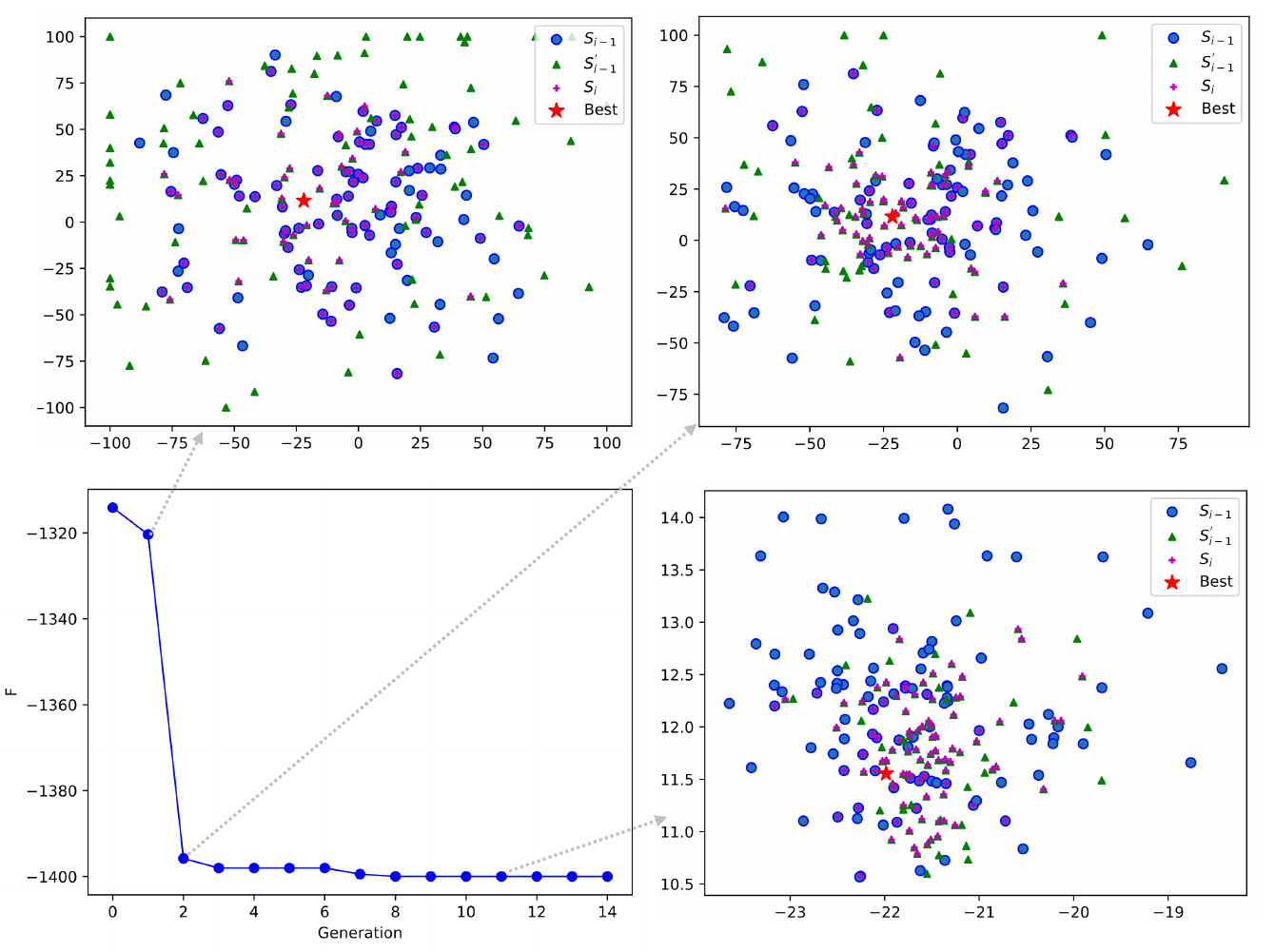}}
\caption{Visualization of the optimization process.}
\label{fig:a5}
\end{center}
%\vspace{-8pt}
\end{figure}

We substantiated the significance of DECN's CRM and SM through visualizing the changes in the population. DECN is comprised of two fundamental functional modules: 1) the generation of potentially valuable solutions (CRM), and 2) the selection of promising solutions (SM). The absence of any of these components would render DECN inoperable. Consequently, conducting ablation experiments by removing specific modules is not feasible. Thus, we take a two-dimensional F4 function as an example to verify that DECN can indeed advance the optimization. In Figure \ref{fig:a5}, as the iteration proceeds, DECN gradually converges. When passing through the first EM module, the CRM is first passed, and the offspring $S_{i-1}^{'}$ are widely distributed in the search space, and the offspring are closer to the optimal solution. Therefore, the CRM generates more potential offspring and is rich in diversity. After the SM update, the generated $S_{i}$ is around the optimal solution, showing that the SM update can keep good solutions and remove poor ones. This phenomenon also leads to a vast improvement after passing through this module. From the population distribution results of the 2nd, 3rd, and 15th EMs, DECN continuously moves the population to the vicinity of the optimal solution.

\section{Conclusions and Discussion}
We successfully designed DECN to learn optimization strategies for optimization automatically. The better performance than SOTA EAs and meta-learning EAs demonstrates that DECN achieves better performance with less computational cost. Moreover, DECN has a strong generalization ability to unseen tasks with different scales and dimensions.

The No Free Lunch theorem underscores the imperative of employing distinct optimization strategies for diverse tasks. This principle constitutes the foundational rationale of the DECN paradigm. DECN, through its parameterized solution generation and selection mechanisms, adeptly leverages task-specific information to adjust its parameters. This dynamic parameter adaptation facilitates a tighter alignment with the objectives of the target task, enabling the rapid amalgamation of individual strengths within a constrained function evaluation budget, ultimately converging towards comparatively superior solutions. Concurrently, the variability among target tasks engenders corresponding variations within the DECN algorithm. In such instances, DECN's optimization strategies may transcend conventional norms, while remaining grounded in reason and tailored to the objectives of the task at hand, yielding favorable outcomes at the very least.

The designed loss function in DECN effectively promotes the exploitation operation but does not strongly encourage exploration of the fitness landscape. Nevertheless, there exist multiple strategies to strike a balance between exploration and exploitation. For instance, we can enhance the loss function by incorporating a constructed Bayesian posterior distribution \cite{cao2020bayesian} concerning the global optimum. Beyond these adjustments in the loss function to bolster exploration, additional modules can be conceived for integration into the EM to assist DECN in escaping local optima. It's important to note that DECN primarily focuses on continuous optimization problems without constraints. When addressing challenges such as combinatorial optimization, constrained optimization, or multi-objective optimization, the adaptation of DECN should align with the specific characteristics of the given problem.

\ifCLASSOPTIONcompsoc
  \section*{Acknowledgments}
\else
  \section*{Acknowledgment}
\fi

This work was supported in part by the National Natural Science Foundation of China under Grant 62206205, in part by the Fundamental Research Funds for the Central Universities under Grant XJS211905, in part by the Guangdong High-level Innovation Research Institution Project under Grant 2021B0909050008, and in part by the Guangzhou Key Research and Development Program under Grant 202206030003.
%\fi

\ifCLASSOPTIONcaptionsoff
  \newpage
\fi

\end{document}